\documentclass{article} % For LaTeX2e
\usepackage[accepted]{icml2024}

\usepackage{tikz}
\usepackage{epsfig}
\usepackage{graphicx}
\usepackage{amsmath}
\usepackage{amssymb}
\usepackage{microtype}
\usepackage{mathrsfs}
\usepackage{subfigure}
\usepackage{booktabs} 
\usepackage{mathtools}
\usepackage{amsthm}
\usepackage{algorithm}
\usepackage{algorithmic}
\usepackage{tabularray}
\usepackage{wrapfig}

\DeclareMathOperator*{\argmax}{arg\,max}

\newtheorem{theorem}{Theorem}[section]

\newtheorem{lemma}[theorem]{Lemma}
\newtheorem{corollary}[theorem]{Corollary}

\usepackage[utf8]{inputenc} % allow utf-8 input
\usepackage[T1]{fontenc}    % use 8-bit T1 fonts
\usepackage[breaklinks=true,colorlinks,bookmarks=false]{hyperref}
\definecolor{mydarkblue}{rgb}{0,0.08,0.45}
\definecolor{mydarkgreen}{RGB}{0, 139, 69}
\hypersetup{
	colorlinks=true,
	urlcolor=magenta,
	citecolor=mydarkblue,
}
\definecolor{mycyan}{cmyk}{.3,0,0,0}
\usepackage{url}            % simple URL typesetting
\usepackage{booktabs}       % professional-quality tables
\usepackage{amsfonts}       % blackboard math symbols
\usepackage{nicefrac}       % compact symbols for 1/2, etc.
\usepackage{microtype}      % microtypography
\usepackage{xcolor}         % colors
\usepackage{wrapfig}
\usepackage{multirow}
\usepackage{bm}
\usepackage{tabu}
\newcommand{\vect}[1]{\boldsymbol{\mathbf{#1}}}

\usepackage[capitalize]{cleveref}
\crefname{section}{Sec.}{Secs.}
\Crefname{section}{Section}{Sections}
\Crefname{table}{Table}{Tables}
\crefname{table}{Table}{Tables}

\begin{document}
\twocolumn[
\icmltitle{Robust Classification via a Single Diffusion Model}

% Authors must not appear in the submitted version. They should be hidden
% as long as the \iclrfinalcopy macro remains commented out below.
% Non-anonymous submissions will be rejected without review.

\begin{icmlauthorlist}
\icmlauthor{Huanran Chen}{aaa,yyy}
\icmlauthor{Yinpeng Dong}{yyy,rrr}
\icmlauthor{Zhengyi Wang}{yyy}
\icmlauthor{Xiao Yang}{yyy}
\icmlauthor{Chengqi Duan}{yyy}
\icmlauthor{Hang Su}{yyy}
\icmlauthor{Jun Zhu}{yyy,rrr}
\end{icmlauthorlist}

\icmlaffiliation{yyy}{Dept. of Comp. Sci. and Tech., Institute for AI, Tsinghua-Bosch Joint ML Center, THBI Lab,
 BNRist Center, Tsinghua University, Beijing, 100084, China}
\icmlaffiliation{aaa}{School of Computer Science, Beijing Institute of Technology}
\icmlaffiliation{rrr}{RealAI}

\icmlcorrespondingauthor{Yinpeng Dong}{dongyinpeng@tsinghua.edu.cn}
\icmlcorrespondingauthor{Jun Zhu}{dcszj@tsinghua.edu.cn}

% \icmlaffiliation{yyy}{\makebox[0cm][l]{Tsinghua University}}
% \icmlaffiliation{aaa}{\makebox[4cm][l]{Beijing Institute of Technology}}

% \icmlcorrespondingauthor{Yinpeng Dong}{\makebox[4cm][l]{dongyinpeng@mail.tsinghua.edu.cn}}
% \icmlcorrespondingauthor{Jun Zhu}{\makebox[3cm][l]{dcszj@mail.tsinghua.edu.cn}}

\newcommand{\fix}{\marginpar{FIX}}
\newcommand{\new}{\marginpar{NEW}}

%\iclrfinalcopy % Uncomment for camera-ready version, but NOT for submission.

\icmlkeywords{Machine Learning, ICML}

\vskip 0.3in
]

\begin{abstract}
Diffusion models have been applied to improve adversarial robustness of image classifiers by purifying the adversarial noises or generating realistic data for adversarial training. However, diffusion-based purification can be evaded by stronger adaptive attacks while adversarial training does not perform well under unseen threats, exhibiting inevitable limitations of these methods. To better harness the expressive power of diffusion models, this paper proposes Robust Diffusion Classifier (RDC), a generative classifier that is constructed from a pre-trained diffusion model to be adversarially robust. RDC first maximizes the data likelihood of a given input and then predicts the class probabilities of the optimized input using the conditional likelihood estimated by the diffusion model through Bayes' theorem. To further reduce the computational cost, we propose a new diffusion backbone called multi-head diffusion and develop efficient sampling strategies. As RDC does not require training on particular adversarial attacks, we demonstrate that it is more generalizable to defend against multiple unseen threats. In particular, RDC  achieves $75.67\%$ robust accuracy against various $\ell_\infty$ norm-bounded adaptive attacks with $\epsilon_\infty=8/255$ on CIFAR-10, surpassing the previous state-of-the-art adversarial training models by $+4.77\%$. The results highlight the potential of generative classifiers by employing pre-trained diffusion models for adversarial robustness compared with the commonly studied discriminative classifiers. Code is available at \url{https://github.com/huanranchen/DiffusionClassifier}.
\end{abstract}

\printAffiliationsAndNotice{}

\section{Introduction}

A longstanding problem of deep learning is the vulnerability to adversarial examples \citep{szegedy2013intriguing, fgsm}, which are maliciously generated by applying human-imperceptible perturbations to natural examples, but can cause deep learning models to make erroneous predictions. As the adversarial robustness problem leads to security threats in real-world applications (e.g., face recognition \citep{Sharif2016Accessorize, dong2019efficient}, autonomous driving \citep{cao2021invisible,jing2021too}, healthcare \citep{finlayson2019adversarial}), there is a lot of work on defending against adversarial examples, such as adversarial training \citep{pgd,zhang2019theoretically,wang2023better_diffusion_improve_AT}, image denoising \citep{liao2018defense,samangouei2018defense,song2017pixeldefend}, certified defenses \citep{raghunathan2018certified,wong2018provable,cohen2019certified}.

%\hangx{may following the logic: discriminative classifier -> generative classifier -> diffusion ?}
%Despite the great success of deep learning, it is vulnerable to adversarial attacks, which impose human-imperceptible perturbations to the natural images~\citep{MI, pgd, chen2023rethinking_model_ensemble}, misleading the deep learning classifiers, posing significant threats to the application of neural networks~\citep{adversarialexampleinphysicalworld, tsea, yang2023towards_attack_3d_face}. Researchers have proposed many defense algorithms to defend against adversarial attacks, greatly improving the security of deep neural networks.

%\hangx{what are the limitations of the previous methods? why diffusion? may introduce the generative classifier here?} 
Recently, diffusion models have emerged as a powerful family of generative models, consisting of a forward diffusion process that gradually perturbs data with Gaussian noise and a reverse generative process that learns to remove noise from the perturbed data \citep{sohl2015deep,ddpm,nichol2021improved,song2020score_diffusion_sde}. Some researchers have tried to apply diffusion models to improving adversarial robustness in different ways. For example, the adversarial images can be purified through the forward and reverse processes of diffusion models before feeding into the classifier \citep{blau2022threat,nie2022diffpure,wang2022guided}. 
Besides, the generated data from diffusion models can significantly improve adversarial training \citep{rebuffi2021fixing_data_aug_improve_at,wang2023better_diffusion_improve_AT}, achieving the state-of-the-art results on robustness benchmarks \citep{croce2020robustbench}. These works show promise of diffusion models in the field of adversarial robustness.

%Recently, diffusion models have dominated the generative task due to their excellent performance~\citep{dhariwal2021diffusion_beat_gan, rombach2022latent_diffusion}. Researchers have tried to introduce diffusion models to adversarial defenses to further improve the adversarial robustness. Some researchers propose to use the diffusion model as a denoising purifier, named DiffPure, which aims to purify the adversarial perturbation before feeding the images into classifier~\citep{nie2022diffpure, carlini2022certified_diffpure_free, xiao2022densepure}. Others use the diffusion models as effective data augmenters to further improve adversarial training~\citep{wang2023better_diffusion_improve_AT}. These methods greatly improve the robust accuracy in adversarial defense leader board~\citep{croce2020robustbench}, demonstrating the effectiveness of diffusion models.

However, the existing methods have some limitations. On one hand, the diffusion-based purification approach incurs much more randomness compared to conventional methods, and can be effectively attacked by using the exact gradient and a proper step size \footnote{We lower the robust accuracy of DiffPure \citep{nie2022diffpure} from 71.29\% to 44.53\% under the $\ell_\infty$ norm with $\epsilon_{\infty}=8/255$, and from 80.60\% to 75.59\% under the $\ell_2$ norm with $\epsilon_{2}=0.5$, as shown in Table~\ref{table:robust_generalization}.}. We observe that the adversarial example cannot make the diffusion model output an image of a different class, but the perturbation is not completely removed. Therefore, the poor robustness of diffusion-based purification is largely due to the vulnerability of downstream classifiers.
On the other hand, although adversarial training methods using data generated by diffusion models achieve excellent performance, they are usually not generalizable across different threats \citep{tramer2019adversarial, nie2022diffpure}. 
In summary, these methods leverage diffusion models to improve adversarial robustness of discriminative classifiers, 
but discriminative learning cannot capture the underlying structure of data distribution,
making it hard to control the predictions of inputs outside the training distribution \citep{schott2019towards}. % still have some drawbacks, which we think could be due to the discriminative learning itself\hangx{what are the limitation, better provide more evidence for discriminative models?}. 
As a generative approach, diffusion models provide a more accurate estimation of score function (i.e., the gradient of log-density at the data point) across the entire data space \citep{song2019generative,song2020score_diffusion_sde}, which also have the potential to provide accurate class probabilities.  
%\hangx{why it can facilate the generative classifier } 
Therefore, we devote to exploring \emph{how to convert a diffusion model into a generative classifier for improved adversarial robustness?}

%sGreat improvement of adversarial training by data augmentation using the diffusion models~\citep{wang2023better_diffusion_improve_AT} also demonstrates the strong effectiveness of diffusion models\yinpeng{what is the limitation of this approach?} \huanran{No limitation.... This is Niubility}. Based on this observation, we question that: Could we classify the images via a single diffusion model? Will the diffusion classifier be robust against adversarial attacks?\yinpeng{any motivation on why diffusion model could be a robust classifier?}\huanran{Explain in the first paragraph in Methodology}

\begin{figure*}
    \centering
   \includegraphics[width=.99\linewidth]{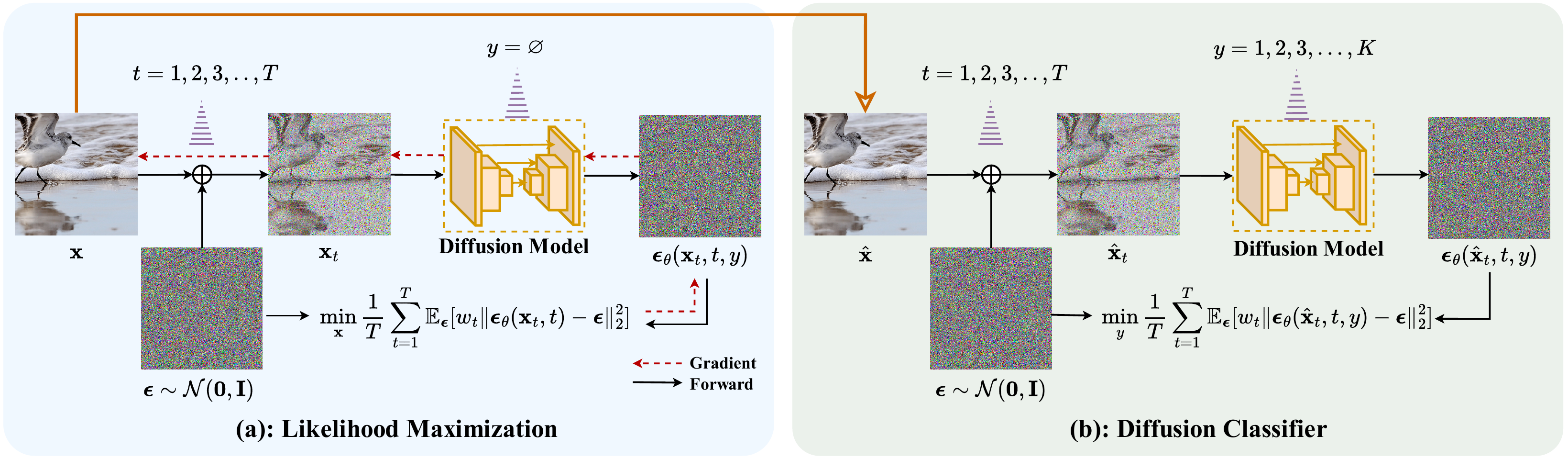}
    \vspace{-1ex}\caption{Illustration of our proposed Robust Diffusion Classifier~(RDC). Given an input image $\vect{x}$, our approach first maximizes the data likelihood (Left) and then classifies the optimized image $\hat{\vect{x}}$ with a diffusion model (Right). The class probability $p(y|\hat{\vect{x}})$ is given by the conditional log-likelihood $\log p_\theta(\hat{\vect{x}}|y)$, which is approximated by the variational lower bound involving calculating the noise prediction error (i.e., diffusion loss) averaged over different timesteps for every class.  
    }
    \label{fig:illustration}
    % \vspace{-2ex}
\end{figure*}

In this paper, we propose Robust Diffusion Classifier (\textbf{RDC}), a generative classifier obtained from a single pre-trained diffusion model to achieve adversarial robustness. Our method calculates the class probability $p(y|\vect{x})$ using the conditional likelihood $p_{\theta}(\vect{x}|y)$ estimated by a diffusion model through Bayes' theorem. The conditional likelihood is approximated by the variational lower bound, which involves calculating the noise prediction loss for every class under different noise levels. In order to reduce time complexity induced by the number of classes, we propose a new UNet backbone named \textbf{multi-head diffusion} by modifying the last convolutional layer to output noise predictions of all classes simultaneously. Theoretically, we validate that the optimal diffusion model can achieve absolute robustness under common threat models.
%To compute $p(y|\vect{x})$ for a given input image, we utilize the diffusion loss to approximate the log-likelihood of this image, then we calculate $p(y|\vect{x})$ by $\log p(\vect{x}|y)$ using Bayes Theorem. 
%Besides, we theoretically analyze the robustness of our diffusion classifier. We derive the optimal solution for the diffusion model and diffusion classifier which achieves absolute robustness under $\el\ell_\infty$ norm of $8/255$ and $\ell_2$ norm of $0.5$. 
However, the practical diffusion model may have an inaccurate density estimation $ p_{\theta}(\vect{x}|y)$ or a large gap between the likelihood and its lower bound, leading to inferior performance. 
To address this issue, we further propose \textbf{Likelihood Maximization} as a pre-optimization step to move the input data to regions of high likelihoods before feeding into the diffusion classifier. 
Our RDC, directly constructed from a pre-trained diffusion model without training on specific adversarial attacks, can perform robust classification under various threat models.
%\hangx{training jointly or separately? } \hangx{more details about why and how we address the previous issues?}
%We demonstrate that the likelihood maximization and diffusion classifier are well coupled, and the  RDC can accurately classify adversarial examples with a single diffusion model.
%We conduct extensive experiments to validate the effectiveness of our method, e.g., on CIFAR-10
%Finally, we analyze the adversarial examples of our diffusion classifier, which are caused by the inaccurate prediction of $\log p(\vect{x}|y)$ or inaccurate approximation of $\log p(\vect{x}|y)$ by the diffusion loss, both of which could be solved by minimizing the diffusion loss.
%Based on these observations, we propose a Robust Diffusion Classifier, which minimize the diffusion loss for input images before feeding them into diffusion classifier, achieving state-of-the-art robustness.\yinpeng{rewrite this paragraph}

We empirically compare our method with various state-of-the-art methods against strong adaptive attacks, which are integrated with AutoAttack \citep{autoattack} for more comprehensive evaluations. Specifically, at each step, we obtain the gradient through adaptive attacks (e.g., BPDA, exact gradient) and then feed the gradient into AutoAttack to perform update. 
Additionally, we investigate the gradient randomness and find that the gradient variance in our method is exceptionally low. This suggests that, due to the low variance and precise gradient, our method does not result in obfuscated gradients \citep{athalye2018obfuscated_gradient}, indicating that the evaluation is accurate and reliable.
On CIFAR-10 \citep{krizhevsky2009learning}, RDC achieves 75.67\%
robust accuracy under the $\ell_\infty$ norm threat model with $\epsilon_\infty=8/255$, exhibiting a $+4.77\%$ improvement over the state-of-the-art adversarial training method~\citep{wang2023better_diffusion_improve_AT}, and a $+3.01\%$ improvement over the state-of-the-art dynamic defenses and randomized defenses \citep{perez2021enhancing,blau2023classifier}.
Under unseen threats, RDC leads to a more significant improvement ($>30\%$) over adversarial training models, DiffPure~\citep{nie2022diffpure} and generative classifiers.
%Even without likelihood maximization, the diffusion classifier itself can achieve $35.94\%$ and $35.94\%$ robust accuracy under the $\ell_\infty$ and $\ell_2$ norm, which is non-trivial 
% We further conduct thorough analysis of more carefully-designed adaptive attacks and various ablation studies to verify that the achieved robustness is not caused by gradient obfuscation \citep{athalye2018obfuscated_gradient}. 
Our results disclose the potential of generative models for solving the adversarial robustness problem.
%, demonstrating the effectiveness of our method. 

%On CIFAR10 dataset, our Diffusion Classifier achieve $35.94\%$ under $\ell_\infty$ norm $8/255$ and $xx.xx\%$ under $\ell_2$ norm $0.5$, surpassing all previous generative classifiers. Besides, our Robust Diffusion Classifier achieve $73.24\%$ under $\ell_\infty$ norm $8/255$ and $x.xx\%$ under $\ell_2$ norm $0.5$, achieve state-of-the-art robustness. More surprisingly, our defenses achieve $xxxxx\%$ against STAdv, demonstrate the strong generalization ability against unseen threat models.
\section{Related work}

% Given a classifier $f: R^\mathcal{D} \to R^n$, where $\mathcal{D}$ is the data dimension and  $n$ is the number of candidate labels, adversarial attack is to add human-imperceptible noise $\vect{\delta}$~(constrained by $||\vect{\delta}|| \leq \epsilon_{adv}$) to an input image $\vect{x}$ with true label $y$, misleading the classifier, i.e. $\arg\max_{\hat{y}} f(\vect{x}+\vect{\delta})_{\hat{y}} \neq y$.

% Gradient-based adversarial attacks are the most effective algorithms, which find the adversarial examples by optimizing surrogate loss function $L(f(\vect{x+\delta}), y)$, such as Cross Entropy Loss, CW Loss~\citep{carlini2017towards}, DLR loss~\citep{autoattack}. These algorithms could achieve nearly 100\% attack success rate against the modern state-of-the-art discriminative classifiers~(without defenses)~\citep{carlini2017towards}, pose a great threat to the application of deep learning.

% Most of adversarial attack algorithms convert \cref{equation:adv_indifferentiable} into a differentiable problem:
% \begin{equation}
%     \arg\min_{\vect{\delta}} L(f(\vect{x+\delta}), y), \;\; s.t. ||\vect{\delta}|| \leq \epsilon_{adv}
% \end{equation}

% Where $L$ is the normal loss function like Cross Entropy Loss, CW Loss~\citep{carlini2017towards}, DLR loss~\citep{autoattack}. 

\textbf{Adversarial robustness.} Adversarial examples \citep{szegedy2013intriguing,fgsm} are widely studied in the literature, which are generated by adding imperceptible perturbations to natural examples, but can mislead deep learning models. Many adversarial attack methods \citep{carlini2017towards,athalye2018obfuscated_gradient, MI,pgd,chen2023rethinking,autoattack} have been proposed to improve the attack success rate under the white-box or black-box settings, which can be used to evaluate model robustness.
%Popular attacks like PGD~\citep{pgd}, AutoAttack~\citep{autoattack} could achieve nearly 100\% attack success rate against the modern state-of-the-art discriminative classifiers~(without defenses)~\citep{carlini2017towards} in the white-box scenario, where the attackers have full information of the attacked models. These attacks encourage researchers to develop more robust and reliable algorithms.
To defend against adversarial attacks, adversarial training \citep{pgd, zhang2019theoretically} stands out as the most effective method, which trains neural networks using adversarially augmented data. However, these models tend to exhibit robustness only to a specific attack they are trained with, and have poor generalization ability to unseen threats \citep{tramer2019adversarial,laidlaw2020perceptual}. Another popular approach is adversarial purification \citep{liao2018defense,samangouei2018defense,song2017pixeldefend,nie2022diffpure}, which denoises the input images for classification. Most of these defenses cause obfuscated gradients~\citep{athalye2018obfuscated_gradient} and can be evaded by adaptive attacks \citep{tramer2020adaptive}. %Recently, researchers also certify the robustness of deep learning models by convex relaxation~\citep{raghunathan2018certified,wong2018provable} or randomized smoothing~\citep{cohen2019certified, carlini2022certified_diffpure_free}. Although these approaches offer provable robustness, they still fall short when compared to the performance of adversarial training.

\textbf{Generative classifiers.}
Generative classifiers, like naive Bayes \citep{ng2001discriminative_vs_generative}, predict the class probabilities $p(y|\vect{x})$ for a given input $\vect{x}$ by modeling the data likelihood $p(\vect{x}|y)$ using generative models. Compared with discriminative classifiers, generative classifiers are often more robust and well-calibrated \citep{ raina2003classification,schott2019towards, li2019_are_generative_classifier_more_robust, mackowiak2021generative}. 
Modern generative models like diffusion models~\citep{ddpm,song2020score_diffusion_sde} and energy-based models~\citep{lecun2006tutorial_ebm,du2019implicit} can also be used as generative classifiers.
SBGC \citep{zimmermann2021score_based_generative_classifier} utilizes a score-based model to calculate the log-likelihood $\log p(\vect{x}|y)$ by integration and calculates $p(y|\vect{x})$ via Bayes' theorem. HybViT \citep{yang2022your_vit_hybride} learns the joint likelihood $\log p(\vect{x}, y)=\log p(\vect{x})+\log p(y|\vect{x})$ by training a diffusion model to learn $\log p(\vect{x})$ and a standard classifier to model $\log p(y|\vect{x})$ at training time, and directly predicts $p(y|\vect{x})$ at test time. JEM \citep{grathwohl2019your_classifier_secret_ebm} utilizes the energy-based model to predict joint likelihood $ \log p(\vect{x}, y)$ and applies Bayes' theorem to get $p(y|\vect{x})$. We also compare with these generative classifiers in experiments. Recently, diffusion models have also been used for generative classification. \citet{hoogeboom2021argmax, han2022card} perform diffusion process in logit space to learn the categorial classification distribution. Concurrent work \citep{clark2023text,li2023your} converts diffusion models to generative classifiers in a similar way to ours, but they focus on zero-shot classification while do not consider adversarial robustness.  
%To our best knowledge, the robustness of our diffusion classifier outperforms all previous generative classifiers.

\textbf{Diffusion models for adversarial robustness.}
As a powerful family of generative models~\citep{dhariwal2021diffusion_beat_gan,rombach2022latent_diffusion}, diffusion models have been introduced to further improve adversarial robustness. DiffPure \citep{nie2022diffpure} utilizes diffusion models to purify adversarial perturbations by first adding Gaussian noise to input images and then denoising the images. Diffusion models can also help to improve the certified robustness with randomized smoothing \citep{carlini2022certified_diffpure_free, xiao2022densepure, zhang2023diffsmooth, chen2024your}. Besides, using data generated by diffusion models can significantly improve the performance of adversarial training \citep{rebuffi2021fixing_data_aug_improve_at,wang2023better_diffusion_improve_AT}. %utilizes a diffusion model as a data augmenter to generate large amounts of data, significantly improving adversarial training. 
%However, there are still some limitations of the existing methods. 
% On the one hand, DiffPure is a kind of stochastic gradient~\citep{athalye2018obfuscated_gradient}, which could be efficiently attacked by using the exact gradient and a proper step size. Fortunately, we observe that the adversarial example cannot make the diffusion model output an image of another class, but the adversarial perturbation for the downstream classifier is not completely removed. Therefore, the poor robustness of diffusion-based purification is mainly due to the vulnerability of downstream classifiers instead of the diffusion model itself. 
% On the other hand, AT-EDM is still vulnerable to unseen threat models due to the drawback of adversarial training~\citep{laidlaw2020perceptual,tramer2019adversarial}. 
% Generally, these methods utilize diffusion models along with discriminative classifiers for improving adversarial robustness, but cannot completely mitigate the non-robustness caused by downstream classifiers.
% Inspired by these phenomenons, we aim to explore the following question: \emph{Can we use a single diffusion model as a generative classifier for improved adversarial robustness?}
However, DiffPure is vulnerable to stronger adaptive attacks while adversarial training models do not generalize well across different threat models, as shown in Table~\ref{table:robust_generalization}.
%On the one hand, there still exist some adversarial examples that diffusion model could not purify.
%On the other hand, AT-EDM is still vulnerable to unseen threat models due to the drawback of adversarial training~\citep{laidlaw2020perceptual,tramer2019adversarial}. 
A potential reason of their problems is that they still rely on discriminative classifiers, which do not capture the underlying structure of data distribution. As diffusion models have more accurate score estimation in the whole data space, we aim to explore whether a diffusion model itself can be leveraged to build a robust classifier.
%Generally, these methods utilize diffusion models along with discriminative classifiers for improving adversarial robustness, but cannot completely mitigate the non-robustness caused by downstream classifiers.
%Inspired by these phenomenons, we aim to explore the following question: \emph{Can we use a single diffusion model as a generative classifier for improved adversarial robustness?}

\section{Methodology}

In this section, we present our \textbf{Robust Diffusion Classifier (RDC)}, a generative classifier constructed from a pre-trained diffusion model. We first provide an overview of diffusion models in \cref{sec:3-1}, then present how to convert a (class-conditional) diffusion model into a diffusion classifier in \cref{sec:3-2} with a robustness analysis considering the optimal setting in \cref{sec:3-2-1}, and finally detail the likelihood maximization and time complexity reduction techniques to further improve the robustness and efficiency in \cref{sec:pre-optimization} and \cref{sec:time_reduction}, respectively. Fig.~\ref{fig:illustration} illustrates our approach.

\subsection{Preliminary: diffusion models}
\label{sec:3-1}

We briefly review discrete-time diffusion models~\citep{ddpm}. Given $\vect{x}:=\vect{x}_0$ from a real data distribution $q(\vect{x}_0)$, the forward diffusion process gradually adds Gaussian noise to the data to obtain a sequence of noisy samples $\{\vect{x}_t\}_{t=1}^{T}$ according to a scaling schedule $\{\alpha_t\}_{t=1}^T$ and a noise schedule $\{\sigma_t\}_{t=1}^T$ as
\begin{equation}
\label{equation:diffusion_forward}
    \begin{aligned}
    q(\vect{x}_t|\vect{x}_0)=\mathcal{N}(\vect{x}_t; \sqrt{\alpha_t} \vect{x}_0, \sigma_t^2 \mathbf{I}).
    \end{aligned}
\end{equation}
Assume that the signal-to-noise ratio $\text{SNR}(t)=\alpha_t/\sigma^2_t$ is strictly monotonically decreasing in time, the sample $\vect{x}_t$ is increasingly noisy during the forward process. The scaling and noise schedules are prescribed such that $\vect{x}_T$ is nearly an isotropic Gaussian distribution.

%To generate images, we need to reverse the diffusion process. It is defined as a Markov chain with learned Gaussian distributions as 
The reverse process for Eq.~\eqref{equation:diffusion_forward} is defined as a Markov chain aimed to approximate $q(\vect{x}_0)$  by gradually denoising from the standard Gaussian distribution $p(\vect{x}_T)=\mathcal{N}(\vect{x}_T;\mathbf{0},\mathbf{I})$: 
\begin{equation}
    \begin{aligned}
        &p_{\theta}(\vect{x}_{0:T}) = p(\vect{x}_T)\prod_{t=1}^Tp_{\theta}(\vect{x}_{t-1}|\vect{x}_t), \\ 
        &p_{\theta}(\vect{x}_{t-1}|\vect{x}_{t})= \mathcal{N}(\vect{x}_{t-1};\vect{\mu}_{\theta}(\vect{x}_t, t), \tilde{\sigma}_t^2\mathbf{I}),
    \end{aligned}
\end{equation}
where $\vect{\mu}_{\theta}$ is generally parameterized by a time-conditioned noise prediction network  $\vect{\epsilon}_{\theta}(\vect{x}_t, t)$  \citep{ddpm, kingma2021variational_diffusion}:
\begin{equation}
    % \vect{\mu}_{\theta}(\vect{x}_t, t) = \sqrt{\frac{\alpha_{t-1}}{\alpha_{t}}}\left(\vect{x}_t-\sqrt{\frac{\sigma_t}{1-\alpha_t}}\vect{\epsilon}_{\theta}(\vect{x}_t, t)\right).
    \vect{\mu}_{\theta}(\vect{x}_t, t) = \sqrt{\frac{\alpha_{t-1}}{\alpha_{t}}}\left(\vect{x}_t-\frac{\sigma_t^2-\frac{\alpha_t}{\alpha_{t-1}}\sigma_{t-1}^2}{\sigma_t}\vect{\epsilon}_{\theta}(\vect{x}_t, t)\right).
\end{equation}
%and learn the time-conditioned noise prediction network $\vect{\epsilon}_{\theta}(\vect{x}_t, t)$.
The reverse process can be learned by optimizing the variational lower bound on log-likelihood as
\begin{align}
\label{equation:elbo_uncondition}
   % \begin{aligned}
        \log p_{\theta}(\vect{x})
        \geq &  \mathbb{E}_{q}[ -D_{\mathrm{KL}}(q(\vect{x}_T|\vect{x}_0)\|p(\vect{x}_T)) + \log p_{\theta}(\vect{x}_0|\vect{x}_1) \nonumber \\
        &- \sum_{t>1} D_{\mathrm{KL}}(q(\vect{x}_{t-1}|\vect{x}_t,\vect{x}_0) \| p_{\theta}(\vect{x}_{t-1}|\vect{x}_t))] \nonumber\\
        %&= \mathbb{E}_q [ - \sum_{t>1}KL(q(x_{t-1}|x_t,x_0) || p_{\theta}(x_{t-1}|x_t))+ C_{un}] \\
        = & - \mathbb{E}_{\vect{\epsilon}, t} \left[ w_t \|\vect{\epsilon}_{\theta}(\vect{x}_t, t) - \vect{\epsilon}\|_2^2 \right] + C_1,
    %\end{aligned}
\end{align}
where $\mathbb{E}_{\vect{\epsilon}, t} [ w_t \|\vect{\epsilon}_{\theta}(\vect{x}_t, t) - \vect{\epsilon}\|_2^2]$ is called the \emph{diffusion loss} \citep{kingma2021variational_diffusion}, $\vect{\epsilon}$ follows the standard Gaussian distribution $\mathcal{N}(\mathbf{0}, \mathbf{I})$, $\vect{x}_t=\sqrt{\alpha_t}\vect{x}_0+\sigma_t\vect{\epsilon}$ given by Eq.~\eqref{equation:diffusion_forward}, $C_{1}$ is typically small and can be dropped \citep{ddpm,song2020score_diffusion_sde}, and $w_t = \frac{\sigma_t\alpha_{t-1}}{2\tilde{\sigma}_t^2(1-\alpha_t)\alpha_t}$. To improve the sample quality in practice, \citet{ddpm} consider a reweighted variant by setting $w_t= 1$. 

Similar to \cref{equation:elbo_uncondition}, the conditional diffusion model $p_{\theta}(\vect{x}|y)$ can be parameterized by $\vect{\epsilon}_{\theta}(\vect{x}_t, t, y)$, while the unconditional model $p_{\theta}(\vect{x})$ can be viewed as a special case with a null input as $\vect{\epsilon}_{\theta}(\vect{x}_t,t)=\vect{\epsilon}_{\theta}(\vect{x}_t, t, y=\varnothing)$. A similar lower bound on the conditional log-likelihood is
\begin{equation}
\label{equation:elbo_condition}
    \log p_{\theta}(\vect{x}|y) \geq - \mathbb{E}_{\vect{\epsilon}, t} \left[ w_t \|\vect{\epsilon}_{\theta}(\vect{x}_t, t, y) - \vect{\epsilon}\|_2^2 \right] + C,
\end{equation}
where $C$ is another negligible small constant.

\subsection{Diffusion model for classification}
\label{sec:3-2}

Given an input $\vect{x}$, a classifier predicts a probability $p_{\theta}(y|\vect{x})$ for class $y\in\{1,2,..,K\}$ over all $K$ classes 
and outputs the most probable class as $\tilde{y}=\argmax_y p_\theta(y|\vect{x})$.
Popular discriminative approaches train Convolutional Neural Networks \citep{alexnet,resnet} or Vision Transformers \citep{vit, liu2021swin} to directly learn the conditional probability $p_{\theta}(y|\vect{x})$. However, these discriminative classifiers cannot predict accurately for adversarial example $\vect{x}^*$ that is close to the real example $\vect{x}$ under the $\ell_p$ norm as $\|\vect{x}^*-\vect{x}\|_p\leq\epsilon_p$, since it is hard to control how inputs are classified outside the training distribution \citep{schott2019towards}.

%To classify input images $\vect{x}$, we first compute the conditional probability $p_{\theta}(y|\vect{x})$ for all possible label $y$, then find the most likely $y$ as our prediction.  
On the other hand, diffusion models are trained to provide accurate density estimation over the entire data space \citep{ddpm, song2019generative,song2020score_diffusion_sde}. By transforming a diffusion model into a generative classifier through Bayes' theorem as $p_{\theta}(y|\vect{x})\propto p_{\theta}(\vect{x}|y)p(y)$, we hypothesize that the classifier can also give a more accurate conditional probability $p_{\theta}(y|\vect{x})$ in the data space, leading to better adversarial robustness. In this paper, we assume a uniform prior $p(y)=1/K$ for simplicity, which is common for most of the datasets \citep{krizhevsky2009learning,russakovsky2015imagenet}.
We show how to compute the conditional probability $p_{\theta}(y|\vect{x})$ via a diffusion model in the following theorem.

\begin{theorem}
\label{theorem:why_optimization} (Proof in \cref{sec:proof_of_diffusion_as_classifier})
Let $d(\vect{x},y, \theta)=\log p_{\theta}(\vect{x}|y) +\mathbb{E}_{\vect{\epsilon}, t}[ w_t\|\vect{\epsilon}_\theta(\vect{x}_t,t, y)-\vect{\epsilon}\|_2^2] $ denote the gap between the log-likelihood and the diffusion loss. 
Assume that $y$ is uniformly distributed as $p(y)=\frac{1}{K}$. If  $d(\vect{x},y,\theta)\to 0$ for all $y$, the conditional probability $p_{\theta}(y|\vect{x})$ is
\begin{equation}
\label{equation:diffusion_classifier}
     p_{\theta}(y|\vect{x}) =  \frac{\exp (-\mathbb{E}_{\vect{\epsilon}, t}[w_t\|\vect{\epsilon}_\theta(\vect{x}_t,t,y)-\vect{\epsilon}\|_2^2])}  {\sum_{\hat{y}}\exp(-\mathbb{E}_{\vect{\epsilon}, t}[w_t\|\vect{\epsilon}_\theta(\vect{x}_t,t,\hat{y})-\vect{\epsilon}\|_2^2])}. 
\end{equation}
\end{theorem}%\junz{a vague statement: how good is the approximation theoretically or empirically or both? It can be very loose if not specified clearly; in other words, the theorem can be meaningless if the approximation is very poor.}

%\begin{proof}
%    (sketch, detail at \cref{sec:proof_of_diffusion_as_classifier}). We first transform the problem of calculating $p_{\theta}(y|\vect{x})$ into the problem of calculating $p_{\theta}(\vect{x}|y)$ through Bayes theorem. Due to we could not directly get the log-likelihood $p_{\theta}(\vect{x}|y)$, we instead use the reconstruction loss~(\cref{equation:elbo_condition}) to estimate the log-likelihood given by diffusion model. To ensure the accuracy of this approximation, we need to force the difference between the lower bound of log-likelihood and log-likelihood to be close to 0, that is $d(\vect{x}, \theta) \to 0$.
%\end{proof}

In \cref{theorem:why_optimization}, we approximate the conditional likelihood with its variational lower bound, which holds true when the gap $d(\vect{x},y,\theta)$ is $0$. In practice, although there is inevitably a gap between the log-likelihood and the diffusion loss, we show that the approximation works well in experiments. %\hangx{in what cases? any requirements?}. 
\cref{equation:diffusion_classifier} requires calculating the noise prediction error over the expectation of random noise $\vect{\epsilon}$ and timestep $t$, which is efficiently estimated with the variance reduction technique introduced in \cref{sec:time_reduction}. Although we assume a uniform prior $p(y)=1/K$ in \cref{theorem:why_optimization}, our method is also applicable for non-uniform priors by adding $\log p(y)$ to the logit of class $y$, where $p(y)$ can be estimated from the training data.
%In practice, for most real examples, we can directly compute $p_{\theta}(y|\vect{x})$ by \cref{equation:diffusion_classifier} for all possible classes $y$ and choose the most likely $y$ as our prediction, without taking $d(x, y, \theta)$ into account.
%Note that our classifier is also applicable in datasets where the prior distribution $p(y)$ is not a uniform distribution if we add the logit of class $y$ by $\log p(y)$. Since the uniform assumption is common in most of the datasets~\citep{krizhevsky2014cifar,russakovsky2015imagenet}, we only focus on this simplified case in this paper. 
Below, we provide an analysis on the adversarial robustness of the diffusion classifier in Eq.~\eqref{equation:diffusion_classifier} under the optimal setting.

%\hangx{whether need to estimate $p(y)$ if $y$ is not uniformly distributed?}

%%%%%%%%%%%%%%%%%%%%%%%%%%%%%%%%%%%%%%%%%%%%%%%%%%%%%%%%%%%%%%%%%%%%%%%%%%%%%%%%%%%%%%%%%%

\subsection{Robustness analysis under the optimal setting}\label{sec:3-2-1}
%\huanran{why not 3.3 instead of 3.2.1}

To provide a deeper understanding of the robustness of our diffusion classifier, we provide a new theoretical result on the optimal solution of the diffusion model (i.e., diffusion model that has minimal diffusion loss over both the training set and the test set), as shown in the following theorem.

%Then we derive the optimal prediction for any class $y$ of our diffusion classifier. Finally, we empirically test the robustness of our diffusion classifier under optimal solution.

\begin{theorem}\label{theorem:3-2}
    (Proof in \cref{sec:optimal_diffusion}) Let $D$ denote a set of examples and $D_y\subset D$ denote a subset whose ground-truth label is $y$. The optimal diffusion model $\vect{\epsilon}_{\theta_D^*}(\vect{x}_t, t, y)$ on the set $D$ is the conditional expectation of \(\vect{\epsilon}\):
\begin{equation}
\label{equation:optimal_diffusion}
    \vect{\epsilon}_{\theta_D^*}(\vect{x}_t, t, y)=\sum_{\vect{x}^{(i)}\in D_y}\frac{1}{\sigma_t} s(\vect{x}_t, \vect{x}^{(i)})\cdot(\vect{x}_t-\sqrt{\alpha_t}\vect{x}^{(i)})
\end{equation}
where $s(\vect{x}_t, \vect{x}^{(i)})$ is the probability that \(\vect{x}_t\) comes from \(\vect{x}^{(i)}\):
% denotes the probability of $\vect{x}_t$ belonging to $\vect{x}^{(i)}$:
\begin{equation*}
    s(\vect{x}_t, \vect{x}^{(i)})=\frac{\exp(-\frac{1}{2\sigma_t^2}\|\vect{x}_t-\sqrt{\alpha_t}\vect{x}^{(i)}\|_2^2)}{\sum_{\vect{x}^{(j)} \in D_y} \exp(-\frac{1}{2\sigma_t^2}\|\vect{x}_t-\sqrt{\alpha_t}\vect{x}^{(j)}\|_2^2)}
\end{equation*}
\end{theorem}

%\begin{proof}
%(sketch, detail at \cref{sec:optimal_diffusion}) Taking the derivative of the training objective $\mathbb{E}_{\vect{\epsilon}, t} \left[ w_t ||\vect{\epsilon}_{\theta}(\vect{x}, t, y) - \vect{\epsilon}||_2^2 \right]$ with respect to $\vect{\epsilon}_{\theta}$, find out the $\vect{\epsilon}_{\theta}$ when the gradient is a zero vector, we can easily get the optimal solution of the diffusion models. 
%\end{proof}

Given the optimal diffusion model in \cref{equation:optimal_diffusion}, we can easily obtain the optimal diffusion classifier by substituting the solution in \cref{equation:optimal_diffusion} into \cref{equation:diffusion_classifier}.

% 这个的intuitive analysis在优化的时候再讲

\begin{corollary} \label{theorem:3-3}
    (Proof in \cref{sec:optimal_diffusion_classifier}) The conditional probability $p_{\theta_D^*}(y|\vect{x})$ given the optimal diffusion model $\vect{\epsilon}_{\theta_D^*}(\vect{x}_t, t, y)$ is \(p_{\theta_D^*}(y|\vect{x}) = \mathrm{softmax}\left(f_{\theta_D^*}(\vect{x})\right)_y\), where
    \begin{equation*}\small
    \begin{aligned}
        &f_{\theta_D^*}(\vect{x})_y=-\mathbb{E}_{\vect{\epsilon}, t}\left[\frac{\alpha_t}{\sigma_t^2}\Big\|\sum_{\vect{x}^{(i)}\in D_y}s(\vect{x}, \vect{x}^{(i)}, \vect{\epsilon}, t)\cdot(\vect{x}-\vect{x}^{(i)})\Big\|_2^2\right], \\
        &s(\vect{x},\vect{x}^{(i)}, \vect{\epsilon}, t)=\frac{\exp\left(-\frac{\|\sqrt{\alpha_t}\vect{x}+\sigma_t\vect{\epsilon}-\sqrt{\alpha_t}\vect{x}^{(i)}\|_2^2}{2\sigma_t^2}\right)}{\sum_{\vect{x}^{(j)}\in D_y}\exp\left(-\frac{\|\sqrt{\alpha_t}\vect{x}+\sigma_t\vect{\epsilon}-\sqrt{\alpha_t}\vect{x}^{(j)}\|_2^2}{2\sigma_t^2}\right)}.
    \end{aligned}
    \end{equation*}
\end{corollary}

%\begin{proof}
%(sketch, detail at \cref{sec:optimal_diffusion_classifier}) Substitute the optimal solution of the diffusion models~(\cref{equation:optimal_diffusion}) into our classifier, we can easily get this result. 
%\end{proof}

\textbf{Remark.} Intuitively, the optimal diffusion classifier utilizes the $\ell_2$ norm of the weighted average difference between the input example $\vect{x}$ and the real examples $\vect{x}^{(i)}$ of class $y$ to calculate the logit for $\vect{x}$. The classifier will predict a label $\tilde{y}$ for an input $\vect{x}$ if it lies more closely to real examples belonging to $D_{\tilde{y}}$. Moreover, the $\ell_2$ norm is averaged with weight $\frac{\alpha_t}{\sigma_t^2}$. As $\frac{\alpha_t}{\sigma_t^2}$ is monotonically decreasing w.r.t. $t$, the classifier gives small weights for noisy examples and large weights for clean examples, which is reasonable since the noisy examples do not play an important role in classification.

% \begin{theorem}
%     The robust radius of our diffusion classifier for a image $x$ with label $y$ is:
% \begin{equation*}
% \centering
% \begin{split}
% &\,\, \,\, \,\, \,\, \,\, \,\, \,\, \,\, \,\, \,\, \,\, \,\, \,\, \,\, \,\, \,\, \,\, \,\, \,\, \,\, \,\, \,\, \,\, \,\, \,\, \,\, \,\, \,\, \,\, \,\, \,\, \,\, \,\, \,\, \,\, \,\, \,\, \,\, \,\, \,\, \,\, \,\, \,\, \,\, \,\, \,\, \,\, \,\, \,\, \,\, \,\, \,\, \,\, \,\, 
% \max_{r} \,\, r\\
% &s.t.\quad  \left\{\begin{array}{lc}
% ||p||<r \\
% \mathbb{E}_{t, \epsilon}\frac{\alpha_t}{\sigma_t^2}[||\sum_{i=1, x_i \in D_j}^{|D_j|}s(x+p, t, i)(x+p-x_i)||_2^2]  \geq \mathbb{E}_{t, \epsilon}\frac{\alpha_t}{\sigma_t^2}[||\sum_{i=1, x_i \in D_y}^{|D_y|}s(x+p, t, i)(x+p-x_i)||_2^2]
% \\
% \end{array}\right.
% \end{split}
% \end{equation*}
% \yinpeng{where is the perturbation $p$ in the second constraint?}
% \end{theorem}

% \begin{proof}
% (sketch, detail at \cref{sec:derivation_robust_radius}) The robust radius $r$ for image $x$ is, any perturbation $p$ in this region, the model will not change the output. That is, the logit of class $y$ is bigger than other classes. Thus, we can get the above result from \cref{equation:diffusion_classifier}.\yinpeng{How to compute the radius? The optimized radius could have error?}
% \end{proof}

Given this theoretical result, we can readily analyze the problem in diffusion models and diffusion classifiers by comparing the optimal solution with the empirical one. We evaluate the robust accuracy of the optimal diffusion classifier under the $\ell_{\infty}$ norm with $\epsilon_{\infty}=8/255$ and the $\ell_2$ norm with $\epsilon_2=0.5$ by AutoAttack \cite{autoattack}. Since our method does not cause obfuscated gradients (as discussed in \cref{sec:adaptive_attack,appendix:more_exp}), the robustness evaluation is accurate. We find that the optimal diffusion classifier achieves 100\% robust accuracy, validating our hypothesis that the accurate density estimation of diffusion models facilitates robust classification. However, the diffusion models are not optimal in practice. Our trained diffusion classifier can only achieve 35.94\% and 76.95\% robust accuracy under the $\ell_{\infty}$ and $\ell_2$ threats, as shown in Table \ref{table:robust_generalization}.
% Despite the non-trivial performance without adversarial training, it still lags behind the state-of-the-art. 

To figure out the problem, we examine the empirical model and the optimal one on adversarial examples. We find that the diffusion loss $\mathbb{E}_{\vect{\epsilon}, t}[ w_t\|\vect{\epsilon}_\theta(\vect{x}_t,t, y)-\vect{\epsilon}\|_2^2]$ of the empirical model is much larger. It is caused by either the inaccurate density estimation of $p_{\theta}(\vect{x}|y)$ of the diffusion model or the large gap between the log-likelihood and the diffusion loss violating $d(\vect{x},y,\theta)\to 0$. Developing a better conditional diffusion model can help to address this issue, but we leave this to future work. In the following section, we propose an optimization-based algorithm as an alternative strategy to solve both problems simultaneously with a pre-trained diffusion model.

%Within nearly all adversarial examples, we observe explosions of the reconstruction loss $\mathbb{E}_{\vect{\epsilon}}[w_t||\vect{\epsilon}_\theta(\vect{x}_t,t,y)-\vect{\epsilon}||_2^2]$. 
%There are only two reasons that could explain this phenomenon. 
%First, our diffusion model gives an inaccurate prediction of $\log_{\theta} p(\vect{x}|y)$. 
%Second, the departure from assumption $d(\vect{x}, \theta) \to 0$, leading to the inaccurate approximation of $\log p_{\theta}(\vect{x}|y)$ by the reconstruction loss $-\mathbb{E}_{\vect{\epsilon}, t}[w_t||\vect{\epsilon}_\theta(\vect{x}_t,t,y)-\vect{\epsilon}||_2^2]$. 
%Both could be solved by minimizing $\mathbb{E}_{\vect{\epsilon}, t}[w_t||\vect{\epsilon}_\theta(\vect{x}_t,t,y)-\vect{\epsilon}||_2^2]$ for any $y$.

% From our point of view, there are two primary factors for this gap. Firstly, the inability of diffusion models to generalize well to the unseen dataset, cause an inaccurate log-likelihood estimation $\log p_{\theta}(x)$, poses challenges in accurately classifying natural examples and reduces robustness to adversarial examples. This necessitates the creation of a better conditional diffusion model, which presents an area for future research. 
% Secondly, within nearly all adversarial examples, we observe  In this paper, we focus our efforts on achieving the assumption that $d(x, \theta) \to 0$ when our diffusion model gives a relative accurate estimation of $\log p_{\theta}(x)$.

%%%%%%%%%%%%%%%%%%%%%%%%%%%%%%%%%%%%%%%%%%%%%%%%%%%%%%%%%%%%%%%%%%%%%%%%%%%%%%%%%%%%%%%%%%%%%%%%

\subsection{Likelihood maximization}
\label{sec:pre-optimization}

To address the above problem, a straightforward approach is to minimize the diffusion loss $\mathbb{E}_{\vect{\epsilon}, t}[ w_t\|\vect{\epsilon}_\theta(\vect{x}_t,t, y)-\vect{\epsilon}\|_2^2]$ w.r.t. $\vect{x}$ such that the input can escape from the region that the pre-trained diffusion model cannot provide an accurate density estimation or the gap between the likelihood and diffusion loss $d(\vect{x}, y, \theta)$ is large. However, we do not know the ground-truth label of $\vect{x}$, making the optimization infeasible. As an alternative strategy, we propose to minimize the unconditional diffusion loss as
%To avoid the input image $\vect{x}$ is in the region that our diffusion classifier could not give an accurate $\log p_{\theta}(\vect{x}|y)$, or do not satisfy $d(\vect{x}, y, \theta) \to 0$, we propose to first minimize unconditional diffusion loss $\mathbb{E}_{\vect{\epsilon}, t}[w_t||\vect{\epsilon}_\theta(\vect{x}_t,t)-\vect{\epsilon}||_2^2]$ before feeding the inputs into the classifier. We name this procedure as Likelihood Maximization, and Diffusion Classifier with Likelihood Maximization as Robust Diffusion Classifier.
\begin{equation}
\label{equation:optimization_real}
    \min_{\hat{\vect{x}}} \mathbb{E}_{\vect{\epsilon}, t}[w_t\|\vect{\epsilon}_\theta(\hat{\vect{x}}_t,t)-\vect{\epsilon}\|_2^2],\quad \text{s.t.}\; \|\hat{\vect{x}}-\vect{x}\|_{\infty} \leq \eta,
\end{equation}
which maximizes the lower bound of the log-likelihood in \cref{equation:elbo_uncondition}, and thus it can not only minimize the gap $d(\vect{x}, \theta)$, but also increase the likelihood \(p(\vect{x})\). We call this approach \textbf{Likelihood Maximization}.
In \cref{equation:optimization_real}, we restrict the $\ell_\infty$ norm between the optimized input $\hat{\vect{x}}$ and the original input $\vect{x}$ to be smaller than $\eta$, in order to avoid optimizing $\hat{\vect{x}}$ into the region of other classes. We solve the problem in \cref{equation:optimization_real} by gradient-based optimization with $N$ steps.

This method can be also viewed as a new diffusion-based purification defense. 
On one hand, \citet{xiao2022densepure} prove that for purification defense, a higher likelihood and a smaller distance to the real data of  the purified input $\hat{\vect{x}}$ tends to result in better robustness. Compared to DiffPure, our method restricts the optimization budget by $\eta$, leading to a smaller distance to the real data. Besides, unlike DiffPure which only maximizes the likelihood with a high probability \citep{xiao2022densepure}, we directly maximize the likelihood, leading to improved robustness.
On the other hand, the adversarial example usually lies in the vicinity of its corresponding real example of the ground-truth class $y$, thus moving along the direction towards higher $\log p(\vect{x})$ will probably lead to higher $\log p(\vect{x}|y)$. Therefore, the optimized input $\hat{\vect{x}}$ could be more accurately classified by the diffusion classifier.

\begin{algorithm}[t] %tb
\small
   \caption{Robust Diffusion Classifier (RDC)}
   \label{algorithm}
\begin{algorithmic}[1]
   \REQUIRE
   A pre-trained diffusion model $\vect{\epsilon}_{\theta}$, input image $\vect{x}$, optimization budget $\eta$, step size $\gamma$,
   optimization steps $N$, momentum decay factor $\mu$.
   \STATE \textbf{Initialize:} $\vect{m}=0, \hat{\vect{x}}=\vect{x}$;
   \FOR{$n=0$ {\bfseries to} $N-1$}
   %\STATE \# first step
   \STATE Estimate $\vect{g}=\nabla_{\vect{x}} \mathbb{E}_{\vect{\epsilon}, t}[w_t\|\vect{\epsilon}_\theta(\hat{\vect{x}}_t,t)-\vect{\epsilon}\|_2^2]$ using one randomly sampled $t$ and $\vect{\epsilon}$;
   \STATE Update momentum $\vect{m} = \mu \cdot \vect{m} - \frac{\vect{g}}{\|\vect{g}\|_1}$;
   \STATE Update $\hat{\vect{x}}$ by $\hat{\vect{x}}=\mathrm{clip}_{\vect{x},\eta}(\hat{\vect{x}}+\gamma \cdot \vect{m})$;
   \ENDFOR
   %\STATE Set $\hat{\vect{x}}$ = $\vect{x}_{N}$;
   %\FOR{$y=1$ {\bfseries to} $K$}
   %\STATE \# first step
   \STATE Calculate $ \mathbb{E}_{\vect{\epsilon}, t}[w_t\|\vect{\epsilon}_\theta(\hat{\vect{x}}_t,t, y)-\vect{\epsilon}\|_2^2]$ for all $y \in \{1,2,...,K\}$ simultaneously using multi-head diffusion;
   %\ENDFOR
   \STATE Calculate $p_{\theta}(y|\vect{x})$ by \cref{equation:diffusion_classifier};
\STATE \textbf{Return:} $\tilde{y} = \argmax_y p_{\theta}(y|\vect{x})$.
\end{algorithmic}
\end{algorithm}

\begin{table*}
\centering
\caption{Clean accuracy~(\%) and robust accuracy~(\%) of different methods against unseen threats.}
\label{table:robust_generalization}
\setlength{\tabcolsep}{5pt}
\begin{tabu}{l|c|ccccc} 
\toprule
\multirow{2}{*}{Method} & \multirow{2}{*}{Architecture} & \multirow{2}{*}{Clean Acc} & \multicolumn{4}{c}{Robust Acc}                               \\
                        &                               &                            & $\ell_\infty$ norm & $\ell_2$ norm & StAdv & Avg             \\ 
\midrule
AT-DDPM-$\ell_\infty$   & WRN70-16                      & 88.87                      & 63.28              & 64.65         & 4.88  & 44.27           \\
AT-DDPM-$\ell_2$        & WRN70-16                      & 93.16                      & 49.41              & 81.05         & 5.27  & 45.24           \\
AT-EDM-$\ell_\infty$    & WRN70-16                      & 93.36                      & 70.90              & 69.73         & 2.93  & 47.85           \\
AT-EDM-$\ell_2$         & WRN70-16                      & 95.90                      & 53.32              & \bf84.77         & 5.08  & 47.72           \\
PAT-self                & AlexNet                       & 75.59                      & 47.07              & 64.06         & 39.65 & 50.26           \\
\hline
DiffPure ($t^*=0.125$)  & UNet+WRN70-16                          & 87.50                      & 40.62              & 75.59         & 12.89 & 43.03           \\
DiffPure ($t^*=0.1$)    & UNet+WRN70-16                          & 90.97                      & 44.53              & 72.65            & 12.89    & 43.35              \\
SBGC                    & UNet                          & 95.04                      & 0.00               & 0.00          & 0.00  & 0.00            \\
HybViT                  & ViT                           & 95.90                      & 0.00               & 0.00          & 0.00  & 0.00            \\
JEM                     & WRN28-10                      & 92.90                      & 8.20               & 26.37         & 0.05  & 11.54            \\
\hline
LM (ours) & UNet+WRN70-16  & 87.89 & 71.68 & 75.00 & 87.50 & \textbf{78.06} \\
DC (ours)               & UNet                          & 93.55                      & 35.94              & 76.95         & \bf93.55 & \textbf{68.81}  \\
RDC (LM+DC) (ours)            & UNet                          & 89.85                      & \bf75.67              & 82.03         & 89.45 & {\color{red}\textbf{82.38}}  \\
\bottomrule
\end{tabu}
\end{table*}

\subsection{Time complexity reduction}
\label{sec:time_reduction}

\textbf{Accelerating diffusion classifier. }
A common practice for estimating the diffusion loss in \cref{equation:diffusion_classifier} is to adopt the Monte Carlo sampling. However, this will lead to a high variance with few samples or high time complexity with many samples. To reduce the variance with affordable computational cost, we directly compute the expectation over $t$ instead of sampling $t$ as
\begin{equation}
\label{equation:monte_carlo_t}\small
\mathbb{E}_{\vect{\epsilon}, t}[w_t\|\vect{\epsilon}_\theta(\vect{x}_t,t,y)-\vect{\epsilon}\|_2^2] = \frac{\sum_{t =1}^T\mathbb{E}_{\vect{\epsilon}}[w_t\|\vect{\epsilon}_\theta(\vect{x}_t,t,y)-\vect{\epsilon}\|_2^2]}{T}.
\end{equation}
\cref{equation:monte_carlo_t} requires to calculate the noise prediction error for all timesteps. For $\vect{\epsilon}$, we still adopt Monte Carlo sampling, but we show that sampling only one $\vect{\epsilon}$ is sufficient to achieve good performance. We can further reduce the number of timesteps by systematic sampling that selects the timesteps at a uniform interval. Although it does not lead to an obvious drop in clean accuracy, it will significantly affect robust accuracy as shown in \cref{sec:4-5}, because the objective is no longer strongly correlated with log-likelihood after reducing the number of timesteps.

With this technique, the diffusion classifier requires $K\times T$ NFEs (Number of Function Evaluations), which limits its applicability to large datasets. It is because current diffusion models are designed for image generation tasks. They can only provide noise prediction for one class at a time. To obtain the predictions of all classes in a single forward pass, we propose to modify the last convolutional layer in the UNet backbone to predict noises for $K$ classes (i.e., $K \times 3$ dimensions) simultaneously. Thus, it only requires $T$ NFEs for a single image. We name this novel diffusion backbone as \textbf{multi-head diffusions}. More details are in \cref{sec:training_details}.

\textbf{Accelerating likelihood maximization.} To further reduce the time complexity of likelihood maximization, for each iteration, instead of calculating the diffusion loss using all timesteps like \cref{equation:monte_carlo_t}, we only uniformly sample a single timestep to approximate the expectation of the diffusion loss. Surprisingly, this modification not only reduces the time complexity of likelihood maximization from $O(N \times T)$ to $O(N)$, but also greatly improves the robustness. This is because this likelihood maximization induces more randomness, thus it is more effective to smooth the local extrema. We provide more in-depth analysis in \cref{appendix:more_exp}.

Given the above techniques, the overall algorithm of RDC is outlined in \cref{algorithm}.

\vspace{-1.5ex}
\section{Experiments}
\label{sec:exp}

In this section, we first provide the experimental settings in Sec.~\ref{sec:4-1}. We then show the effectiveness of our method compared with the state-of-the-art methods in Sec.~\ref{sec:4-2} and the generalizability across different threat models in Sec.~\ref{sec:4-3}. We provide thorough analysis to examine gradient obfuscation in Sec.~\ref{sec:adaptive_attack} and various ablation studies in Sec.~\ref{sec:4-5}.

%\hangx{if possible, find some method or metric to show that our method is obviously better than the competitors.}
\subsection{Experimental settings}\label{sec:4-1}
\textbf{Datasets and training details.} Following \citet{nie2022diffpure}, we randomly select 512 images from the CIFAR-10 test set \citep{krizhevsky2009learning} for evaluation due to the high computational cost of AutoAttack.  We also conduct experiments on other datasets and other settings in \cref{appendix:more_exp}. We adopt off-the-shelf conditional diffusion model in \citet{karras2022elucidating} and train our multi-head diffusion as detailed in \cref{appendix:supplementary_exps_and_training} for 100 epochs on CIFAR-10 training set.

\textbf{Hyperparameters.} In likelihood maximization, we set the optimization steps $N=5$, momentum decay factor $\mu=1$, optimization budget $\eta=8/255$ (see \cref{sec:4-5} for an ablation study), step size $\gamma=0.1$. For each timestep, we only sample one $\vect{\epsilon}$ to estimate $\mathbb{E}_{\vect{\epsilon}}[w_t\|\vect{\epsilon}_\theta(\vect{x}_t,t,y)-\vect{\epsilon}\|_2^2]$.

\textbf{Robustness evaluation.} Following \citet{nie2022diffpure}, we evaluate the clean accuracy and robust accuracy using AutoAttack~\citep{autoattack} under both $\ell_\infty$ norm of $\epsilon_\infty=8/255$ and $\ell_2$ norm of $\epsilon_2=0.5$. To demonstrate the generalization ability towards unseen threat models, we also evaluate the robustness against StAdv~\citep{stadv} with $100$ steps under the bound of $0.05$. Since computing the gradient through likelihood maximization requires calculating the second-order derivative, we use BPDA \citep{athalye2018obfuscated_gradient} as the default adaptive attack, approximating the gradient with an identity mapping. We conduct more comprehensive evaluations of gradient obfuscation in \cref{sec:adaptive_attack}, where we show that BPDA is as strong as computing the exact gradient. It is important to note that, except for adaptive attacks on DiffPure \citep{nie2022diffpure}, all other attacks solely modify the back-propagation (e.g., BPDA, exact gradient) or the loss function. The iterative updates are all performed by AutoAttack~\citep{autoattack}.

One of our follow-up works also provides a thorough theoretical analysis of the certified robustness of our proposed diffusion classifier. For more details, see \citet{chen2024your}.

%%%%%%%%%%%%%%%%%%%%%%%%%%%%%%%%%%%%%%%%%%%%%%%%%%%%%%%%%%%%%%%%%%%%%%%%%%%%%%%%%%%%%%%%%%%%%%5

\subsection{Comparison with the state-of-the-art}\label{sec:4-2}

We compare our method with the state-of-the-art defense methods, including adversarial training with DDPM generated data (AT-DDPM) \citep{rebuffi2021fixing_data_aug_improve_at}, with EDM generated data (AT-EDM) \citep{wang2023better_diffusion_improve_AT}, and DiffPure~\citep{nie2022diffpure}. We also compare with perceptual adversarial training (PAT-self) \citep{laidlaw2020perceptual} and other generative classifiers, including SBGC \citep{zimmermann2021score_based_generative_classifier}, HybViT \citep{yang2022your_vit_hybride}, and JEM~\citep{grathwohl2019your_classifier_secret_ebm}. Notably, robust accuracy of most baselines does not change much on our selected subset. Additionally, we compare the time complexity and robustness of our model with more methods in \cref{table:cifar_supplementary} in \cref{appendix:more_exp}. % compared with the results on RobustBench~\citep{croce2020robustbench}.

DiffPure incurs significant memory usage and substantial randomness, posing challenges for robustness evaluation. Their proposed adjoint method \citep{nie2022diffpure} is insufficient to measure the model robustness. To mitigate this issue, we employ gradient checkpoints to compute the exact gradient and leverage Expectation Over Time (EOT) to reduce the impact of randomness during optimization. Rather than using the 640 times EOT recommended in \cref{fig:randomness}, we adopt PGD-200~\citep{pgd} with 10 times EOT and a large step size $1/255$ to efficiently evaluate DiffPure.

\cref{table:robust_generalization} shows the results of Likelihood Maximization (LM), Diffusion Classifier (DC) and Robust Diffusion Classifier (RDC) compared with baselines under the $\ell_\infty$ and $\ell_2$ norm threat models. We can see that the robustness of DC outperforms all previous generative classifiers by a large margin. Specifically, DC improves the robust accuracy over JEM by +27.74\% under the $\ell_\infty$ norm and +50.58\% under the $\ell_2$ norm. RDC can further improve the performance over DC, which achieves 75.67\% and 82.03\% robust accuracy under the two settings. Notably, RDC  outperforms the previous state-of-the-art model AT-EDM~\citep{wang2023better_diffusion_improve_AT} by +4.77\% under the $\ell_\infty$ norm. 

% \huanran{Check this }Although RDC does not outperform the best adversarial training models under the $\ell_2$ norm because likelihood maximization adopts the $\ell_\infty$ norm in \cref{equation:optimization_real}, it could be solved by adjusting the optimization constraint (See \cref{appendix:more_exp} for details). In the following section, we demonstrate that using the same optimization constraint allows our method to be threat model agnostic, resulting in strong generalization abilities against unseen threat models.

%%%%%%%%%%%%%%%%%%%%%%%%%%%%%%%%%%%%%%%%%%%%%%%%%%%%%%%%%%%%%%%%%%%%%%%%%%%%%%%%%%%%%%%%%%%%%%%%%%%%%

\subsection{Defense against unseen threats}\label{sec:4-3}

Adversarial training methods often suffer from poor generalization across different threat models, while DiffPure requires adjusting purification noise scales for different threat models, which limits their applicability in real-world scenarios where the threat models are unknown. In contrast, our proposed methods are agnostic to specific threat models. To demonstrates the strong generalization ability of our methods across different threat models, we evaluate the generalization performance of our proposed method by testing against different threats, including $\ell_\infty$, $\ell_2$, and StAdv. 

Table \ref{table:robust_generalization} presents the results, demonstrating that the average robustness of our methods surpasses the baselines by more than 30\%. Specifically, RDC outperforms $\ell_\infty$ adversarial training models by +12.30\% under the $\ell_2$ norm and $\ell_2$ adversarial training models by +22.35\% under the $\ell_\infty$ norm. Impressively, LM, DC and RDC achieve 87.50\%, 93.55\% and 89.45\% robustness under StAdv, surpassing previous methods by more than 53.90\%. These results indicate the strong generalization ability of our method and its potential to be applied in real-world scenarios under unknown threats.

% \input{tables/generalization}

%%%%%%%%%%%%%%%%%%%%%%%%%%%%%%%%%%%%%%%%%%%%%%%%%%%%%%%%%%%%%%%%%%%%%%%%%%%%%%%%%%%%%%%%%%%%%%%%%%%%%%%%%%%%%%%%%%%%%%%%%%%%%%%%%%%%%%%%%%%%%%%%%%%%%%%%%%%%%%%

\subsection{Evaluation of gradient obfuscation}
\label{sec:adaptive_attack}

The Diffusion Classifier (DC) can be directly evaluated using AutoAttack. However, the Robust Diffusion Classifier (RDC) cannot be directly assessed with AutoAttack since attacking Likelihood Maximization (LM) requires calculating the second-order derivative of the diffusion loss.
%the non-differentiable nature of the likelihood maximization. 
In this section, we analyze both gradient randomness and gradient magnitude (in \cref{appendix:more_exp}), demonstrating that \textbf{\emph{our method does not result in gradient obfuscation}}. Furthermore, we establish that \textbf{\emph{our method attains nearly identical robustness under both exact gradient attack and BPDA attack}}. These findings compellingly affirm that our method is genuinely robust rather than being overestimated by (potentially) insufficient evaluations.

% \begin{wraptable}{r}{6cm}\vspace{-4ex}
% \caption{Robust accuracy~(\%) of RDC under different adaptive attacks. BPDA ($N=5$) and Lagrange ($N=5$) are adaptive attacks for RDC with 5 steps of LM, while Exact Gradient ($N=1$) and BPDA ($N=1$) are adaptive attacks for RDC with 1 step.}
% \vspace{0.5ex}
% \begin{tabular}{c|c} 
% \toprule
% Attack & Robust Acc \\
% \midrule
% BPDA ($N=5$) & 75.67 \\
% Lagrange ($N=5$) & 77.54 \\
% Exact Gradient ($N=1$) & 69.53 \\
% BPDA ($N=1$) & 69.92 \\
% \bottomrule
% \end{tabular}
% \label{table:adaptive_attack}
% \vspace{-2ex}
% \end{wraptable}

\begin{table}
\caption{Robust accuracy~(\%) of RDC under different adaptive attacks. Note that \(N\) is the number of optimization steps in Likelihood Maximization (LM), \textbf{not} the number of attack iterations. All iterative updates during these attacks are consistently conducted by AutoAttack.}
\setlength{\tabcolsep}{3pt}
\begin{tabular}{cc|cc} 
\toprule
LM steps ($N$)  & Attack &  Clean Acc & Robust Acc \\
\midrule
5 & BPDA  & 89.85 &75.67 \\
5 & Lagrange  & 89.85 &77.54 \\
1 & Exact Gradient  & 90.71 &69.53 \\
1 & BPDA & 90.71 &69.92 \\
\bottomrule
\end{tabular}
\label{table:adaptive_attack}
\vspace{-2ex}
\end{table}

\begin{figure*}[t]
\centering
\subfigure[Randomness]{
\includegraphics[width=5.5cm]{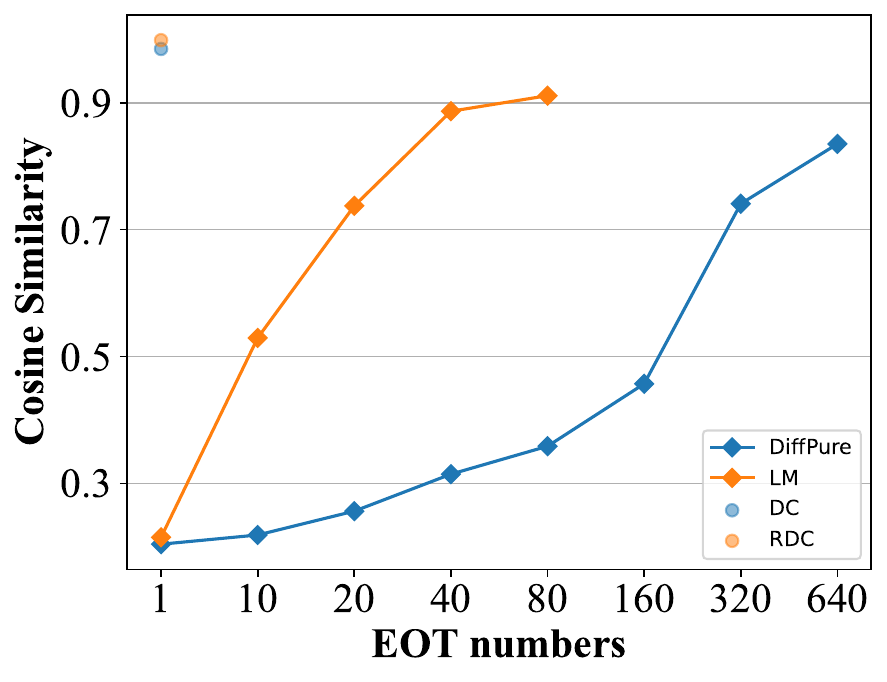}\hspace{-1ex}
\label{fig:randomness}
}
\subfigure[Ablation of $\eta$]{
\includegraphics[width=5.5cm]{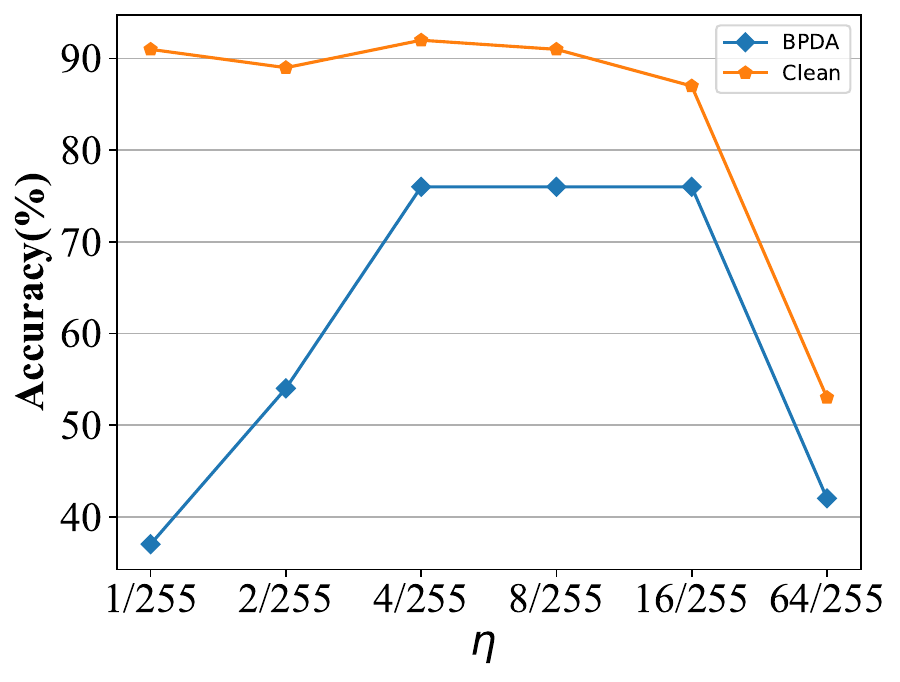}\hspace{-1ex}
\label{fig:ablation_epsilon_o}
}
\subfigure[$T'$]{
\includegraphics[width=5.5cm]{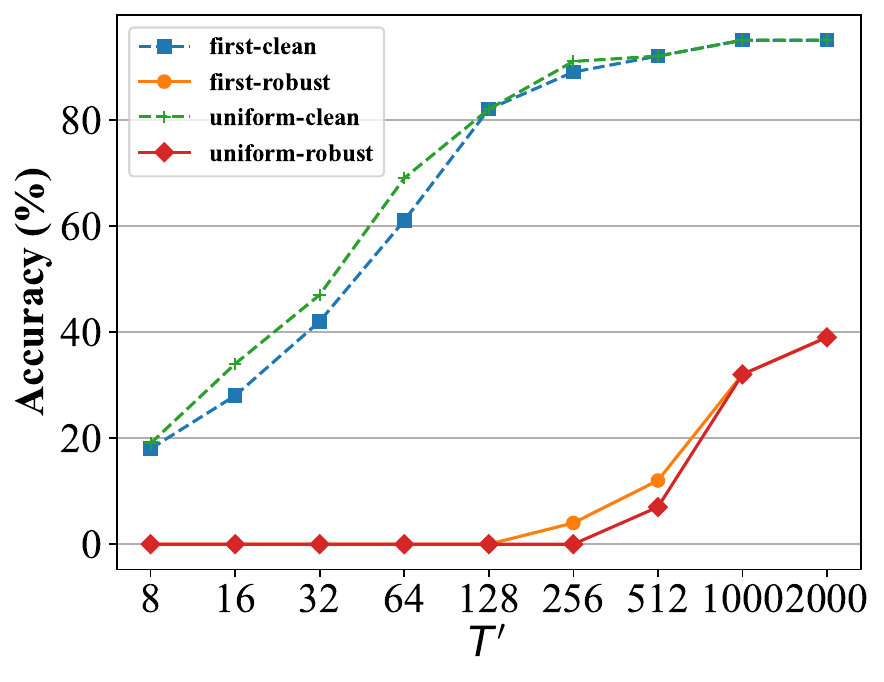}
\label{fig:ablation_ts}
}
\caption{(a): Randomness of different methods. (b-c): Ablation studies of $\eta$ and $T'$.}
\vspace{-1ex}
\end{figure*}

\textbf{Exact gradient attack.} 
To directly evaluate the robustness of RDC, we utilize gradient checkpoint and create a computational graph during backpropagation to obtain exact gradients. However, we could only evaluate RDC when the number of optimization steps $N$ in Likelihood Maximization (LM) is $N=1$ (Note that this is not the number of attack iterations) due to the large memory cost for computing the second-order derivative.
%and technical problem~(i.e., Pytorch~\citep{pytorch} does not support taking the gradient with respect to a non-leaf variable when using gradient checkpoint). 
As shown in \cref{table:adaptive_attack}, our RDC with $N=1$ achieves $69.53\%$ robust accuracy under the exact gradient attack, about $0.39\%$ lower than BPDA. This result suggests that BPDA suffices for evaluating RDC.

\textbf{Lagrange attack.}
RDC optimizes the unconditional diffusion loss before feeding the inputs into DC. If our adversarial examples already have a small unconditional diffusion loss or a large $\log p(\vect{x})$, it may not be interrupted during Likelihood Maximization (LM). Therefore, to produce adversarial examples with a small diffusion loss, we set our loss function as
\begin{equation}
\vspace{-1ex}
\label{equation:lagrange_adaptive}
    \log p_{\theta}(y|\vect{x}) + l \cdot \mathbb{E}_{\vect{\epsilon}, t}[w_t\|\vect{\epsilon}_\theta(\vect{x}_t,t)-\vect{\epsilon}\|_2^2],
\end{equation}
where $p_{\theta}(y|\vect{x})$ is given by \cref{equation:diffusion_classifier}, and the first term is the (negative) cross-entropy loss to induce misclassification.
For an input, we craft adversarial examples using three different weights, $l=0, 1, 10$. If one of these three loss functions successfully generates an adversarial example, we count it as a successful attack. As shown in \cref{table:adaptive_attack}, this adaptive attack is no more effective than BPDA.

\textbf{Gradient randomness.}
To quantify the randomness, we compute the gradients of each model w.r.t. the input ten times and compute the pairwise cosine similarity between the gradients. We then average these cosine similarities across 100 images. To capture the randomness when using EOT, we treat the gradients obtained after applying EOT as a single time and repeat the same process to compute their cosine similarity. As shown in \cref{fig:randomness}, the gradients of our methods exhibit low randomness, while DiffPure is more than 640 times as random as DC, RDC, and about 16 times as random as LM. Thus, the robustness of our methods is not primarily due to the stochasticity of gradients.
% As shown in \cref{equation:cosine_similarity}, for any input pairs $(\vect{x}, y)$, we calculate the gradient $\frac{\partial f(\vect{x})_y}{\partial \vect{x}}$ for k times, and calculate the cosine similarity. Finally, we average the cosine similarity over different input images $\vect{x}$ and candidate label $y$.

% \begin{equation}
%     \mathbb{E}_{\vect{x} \in D} [ \frac{1}{n}\sum_{y=1}^n sim(\{\frac{\partial f_{\theta}(\vect{x})_y}{\partial \vect{x}} \}_{i=1}^{k}) ]
% \label{equation:cosine_similarity}
% \end{equation}

% 记得直接lagrange打diffusion classifier，权重越大攻击成功率越小的奇妙现象

%%%%%%%%%%%%%%%%%%%%%%%%%%%%%%%%%%%%%%%%%%%%%%%%%%%%%%%%%%%%%%%%%%%%%%%%%%%%%%
\subsection{Ablation studies}\label{sec:4-5}
In this section, we perform ablation studies of several hyperparameters with the first 100 examples in the CIFAR-10 test set. All the experiments are done under AutoAttack with BPDA of $\ell_\infty$ bounded perturbations with $\epsilon_\infty = 8/255$.

\textbf{Optimization budget $\eta$.} To find the best optimization budget $\eta$, we test the robust accuracy of different optimization budgets. As shown in \cref{fig:ablation_epsilon_o}, the robust accuracy first increases and then decreases as $\eta$ becomes larger. When $\eta$ is small, we could not move $\vect{x}$ out of the adversarial region. However, when $\eta$ is too large, we may optimize $\vect{x}$ into an image of another class. Therefore, we should choose an appropriate $\eta$. In this work, we set $\eta = 8/255$.

\textbf{Sampling timesteps.} We also attempt to reduce the number of timesteps used in calculating the diffusion loss. Since only the DC is influenced by this parameter, we conduct this experiment exclusively on DC to minimize the impact of other factors. One way is to only calculate the diffusion loss of the first $T'$ timesteps $\{i\}_{i=1}^{T'}$~(``first-clean'' and ``first-robust'' in \cref{fig:ablation_ts}). Inspired by \citet{ddim}, another way is to use systematic sampling, where we use timesteps $\{iT/T'\}_{i=1}^{T'}$~(``uniform-clean'' and ``uniform-robust'' in \cref{fig:ablation_ts} ). Both methods achieve similar results on clean accuracy and robust accuracy. Although a significant reduction of $T'$ does not lead to an obvious drop in clean accuracy, it will significantly affect robust accuracy due to the reason discussed in \cref{sec:time_reduction}.

\textbf{Sampling steps for $\vect{\epsilon}$.} We also attempt to improve the estimation of $\mathbb{E}_{\vect{\epsilon}}[w_t\|\vect{\epsilon}_\theta(\vect{x}_t,t,y)-\vect{\epsilon}\|_2^2]$ by sampling $\vect{\epsilon}$ multiple times or keeping $\vect{\epsilon}$ the same for different timesteps or different classes. However, these increase neither robustness nor accuracy because we have already computed $T$ times for the expectation over $t$. From another perspective, the cosine similarity of the gradients is about 98.48\%, suggesting that additional sampling of $\vect{\epsilon}$ or using the same $\vect{\epsilon}$ is unnecessary.

%%%%%%%%%%%%%%%%%%%%%%%%%%%%%%%%%%%%%%%%%%%%%%%%%%%%%%%%%%%%%%%%%%%%%%%%%%

% \subsubsection{Generating Images by the Gradient from Diffusion Classifier}

% \huanran{Cannot generate images. The reason may be very complicate.}

\section{Conclusion}
In this paper, we propose a novel defense method called Robust Diffusion Classifier (RDC), which leverages a single diffusion model to directly classify input images by predicting data likelihood by diffusion model and calculating class probabilities through Bayes' theorem. We theoretically analyze the robustness of our diffusion classifier, propose to maximize the log-likelihood before feeding the input images into the diffusion classifier. We also propose multi-head diffusion which greatly reduces the time complexity of RDC. We evaluate our method with strong adaptive attacks and conduct extensive experiments. Our method achieves state-of-the-art robustness against these strong adaptive attacks and generalizes well to unseen threat models.

\section*{Impact Statement}

The emergence of adversarial threats in machine learning, especially in critical areas such as autonomous vehicles, healthcare, and financial systems, calls for more robust defense mechanisms. Our work introduces the Robust Diffusion Classifier, a novel framework that harnesses the capabilities of diffusion models for adversarial robustness in image classification. This approach not only contributes to reinforcing the security of machine learning models against adversarial attacks but also demonstrates significant potential in leveraging diffusion models for adversarial robustness. 
Our findings set a new precedent in the field and could be inspireful in enhancing trust in AI applications. While our work primarily focuses on adversarial robustness, it opens avenues for further research for diffusion models in reliability and resilience, moving towards a future where machine learning can reliably function even in adversarially challenging environments.

\section*{Acknowledgements}

This work was supported by the National Natural Science Foundation of China (Nos.  62276149, U2341228, 62061136001, 62076147), BNRist (BNR2022RC01006), Tsinghua Institute for Guo Qiang, and the High Performance Computing Center, Tsinghua University. Y. Dong was also supported by the China National Postdoctoral Program for Innovative Talents. J. Zhu was also supported by the XPlorer Prize.

\bibliography{reference}
\bibliographystyle{icml2024}

\newpage
\appendix
\onecolumn

%%%%%%%%%%%%%%%%%%%%%%%%%%%%%%%%%%%%%%%%%%%%%%%%%%%%%%%%%%%%%%%%%%%%%%%%%%%%%%%%%%%%%%%%%%%%%%%%%%%%%%%%%%%%%%%%%%%%%%%%%%%%%%%%%%%%%%%%%%%%%%%%%%%%%%%%%

\section{Proofs and Derivations}

\subsection{Proof of \cref{theorem:why_optimization}}
\label{sec:proof_of_diffusion_as_classifier}

\begin{proof}
\begin{equation*}
    \begin{aligned}
    p_{\theta}(y|\vect{x})&=\frac{p_{\theta}(\vect{x},y)}{\sum_{\hat{y}}p_{\theta}(\vect{x},\hat{y})} \\
&= \frac{p_{\theta}(\vect{x}|y)p_{\theta}(y)}{\sum_{y}p_{\theta}(\vect{x}|\hat{y})p_{\theta}(\hat{y})} \\
&=\frac{p_{\theta}(\vect{x}|y)}{\sum_{\hat{y}}p_{\theta}(\vect{x}|\hat{y})} \\
&= \frac{e^{\log p_{\theta}(\vect{x}|y)}}{\sum_{\hat{y}}e^{\log p_{\theta}(\vect{x}|\hat{y})}} \\
&= \frac{ \exp \Big( \mathbb{E}_{\vect{\epsilon}, t}[w_t\|\vect{\epsilon}_\theta(\vect{x}_t,t,y)-\vect{\epsilon}\|_2^2]+ \log p_{\theta}(\vect{x}|y)  -\mathbb{E}_{\vect{\epsilon}, t}[w_t\|\vect{\epsilon}_\theta(\vect{x}_t,t,y)-\vect{\epsilon}\|_2^2] \Big) }  
{\sum_{\hat{y}} \exp \Big(\mathbb{E}_{\vect{\epsilon}, t}[w_t\|\vect{\epsilon}_\theta(\vect{x}_t,t,\hat{y})-\vect{\epsilon}\|_2^2]+  \log p_{\theta}(\vect{x}|\hat{y})  -\mathbb{E}_{\vect{\epsilon}, t}[w_t\|\vect{\epsilon}_\theta(\vect{x}_t,t,\hat{y})-\vect{\epsilon}\|_2^2] \Big)  } \\
&= \frac{  \exp \Big( d(\vect{x}, y, \theta) \Big)   \exp \Big(-\mathbb{E}_{\vect{\epsilon}, t}[w\|\vect{\epsilon}_\theta(\vect{x}_t,t,y)-\vect{\epsilon}\|_2^2 ]\Big) } 
{\sum_{\hat{y}} \exp \Big( d(\vect{x}, \hat{y}, \theta) \Big)
\exp \Big( -\mathbb{E}_{\vect{\epsilon}, t}[w_t\|\vect{\epsilon}_\theta(\vect{x}_t,t,\hat{y})-\vect{\epsilon}\|_2^2 ]\Big) } .
    \end{aligned}
\end{equation*}

When $\forall \hat{y}, \; d(\vect{x}, \hat{y}, \theta) \to 0$, we can get:
\begin{equation*}
    \begin{aligned}
        \forall \hat{y}, \,\; \exp \Big(\mathbb{E}_{\vect{\epsilon}, t}[w_t\|\vect{\epsilon}_\theta(\vect{x}_t,t,\hat{y})-\vect{\epsilon}\|_2^2]+  \log p_{\theta}(\vect{x}|\hat{y}) \Big) \to 1.
    \end{aligned}
\end{equation*}

Therefore, 

\begin{equation*}
    p_{\theta}(y|\vect{x}) =  \frac{\exp \Big(-\mathbb{E}_{\vect{\epsilon}, t}[w_t\|\vect{\epsilon}_\theta(\vect{x}_t,t,y)-\vect{\epsilon}\|_2^2] \Big)}  {\sum_{\hat{y}}\exp \Big(-\mathbb{E}_{\vect{\epsilon}, t}[w_t\|\vect{\epsilon}_\theta(\vect{x}_t,t,\hat{y}) -\vect{\epsilon}\|_2^2] \Big)}. 
\end{equation*}
\end{proof}

%%%%%%%%%%%%%%%%%%%%%%%%%%%%%%%%%%%%%%%%%%%%%%%%%%%%%%%%%%%%%%%%%%%%%%%%%%%%%%%%%%%%%%%%%%%%%%%%%%%%%%%%%%%%%%%%%%%%%%%%%%%%%%%%%%%%%%%%%%%%%%%%%%%%%%%%%%%%

\subsection{Derivation of the optimal diffusion classifier}

\subsubsection{Optimal diffusion model: proof of \cref{theorem:3-2}}
\label{sec:optimal_diffusion}

\begin{proof}
The optimal diffusion model has the minimal error $\mathbb{E}_{\vect{x}, t, y} [\|\vect{\epsilon}(\vect{x}_t, t, y) - \vect{\epsilon}\|_2^2]$ among all the models in hypothesis space. Since the prediction for one input pair $(\vect{x}_t, t, y)$ does not affect the prediction for any other input pairs, the optimal diffusion model will give the optimal solution for any input pair $(\vect{x}_t, t, y)$:
\begin{equation*}
    \mathbb{E}_{\vect{x}^{(i)} \sim p(\vect{x}^{(i)}|\vect{x}_t, y)} [\|\vect{\epsilon}_{\theta_D^*}(\vect{x}_t, t, y) - \vect{\epsilon}_i\|_2^2]
    = \min_{\theta} \mathbb{E}_{\vect{x}^{(i)} \sim p(\vect{x}^{(i)}|\vect{x}_t, y)} [\|\vect{\epsilon}_{\theta}(\vect{x}_t, t, y) - \vect{\epsilon}_i\|_2^2],
\end{equation*}
where $\vect{\epsilon}_i = \frac{\vect{x}_t-\sqrt{\alpha_t} \vect{x}^{(i)}}{\sigma_t}$. 

Note that
\begin{equation*}
    p(\vect{x}^{(i)}|\vect{x}_t, y) = 
    \frac{p(\vect{x}^{(i)}| y) p(\vect{x}_t|\vect{x}^{(i)}, y)}{p(\vect{x}_t| y)}
    = 
    \frac{p(\vect{x}^{(i)}| y) q(\vect{x}_t|\vect{x}^{(i)})}{p(\vect{x}_t| y)}.
\end{equation*}

Assume that
\begin{equation*}
\begin{aligned}
p(\vect{x}^{(i)}|y)= \left \{
\begin{array}{ll}
    \frac{1}{|D_y|}                    & ,\vect{x}^{(i)} \in D_y\\
    0     & ,\vect{x}^{(i)} \notin D_y
\end{array}
\right.
\end{aligned}
\end{equation*}

Solving $\frac{\partial }{\partial \vect{\epsilon}_{\theta}(\vect{x}_t, t, y)} \mathbb{E}_{\vect{x}^{(i)} \sim p(\vect{x}^{(i)}|\vect{x}_t, y)} [\|\vect{\epsilon}_{\theta}(\vect{x}_t, t, y) - \vect{\epsilon}_i\|_2^2] = 0$, we can get:

\begin{equation*}
    \begin{aligned}
        &\mathbb{E}_{\vect{x}^{(i)} \sim p(\vect{x}^{(i)}|\vect{x}_t, y)} [\vect{\epsilon}_{\theta}(\vect{x}_t, t, y) - \vect{\epsilon}_i] = 0, \\
        &\sum_{x^{(i)} \in D} p(\vect{x}^{(i)}|\vect{x}_t, y) \vect{\epsilon}_{\theta}(\vect{x}_t, t, y)=\sum_{\vect{x}^{(i)}\in D_y}p(\vect{x}^{(i)}|\vect{x}_t, y)\vect{\epsilon}_i.
    \end{aligned}
\end{equation*}
Substitute $\vect{\epsilon}_i$ by \cref{equation:diffusion_forward}:
\begin{equation*}
    \begin{aligned}
        \vect{\epsilon}_{\theta}(\vect{x}_t, t, y) &= \sum_{\vect{x}^{(i)}\in D_y} p(\vect{x}^{(i)}|\vect{x}_t, y) \frac{\vect{x}_t-\sqrt{\alpha_t} \vect{x}^{(i)}}{\sigma_t} \\
        &= \sum_{\vect{x}^{(i)}\in D_y} \frac{p(\vect{x}^{(i)}| y) q(\vect{x}_t|\vect{x}^{(i)})}{p(\vect{x}_t| y)} \frac{\vect{x}_t-\sqrt{\alpha_t} \vect{x}^{(i)}}{\sigma_t} \\
        &= \sum_{\vect{x}^{(i)}\in D_y} \frac{p(\vect{x}^{(i)}| y)}{p(\vect{x}_t| y)}p(\mathcal{N}(\vect{x}_t|\sqrt{\alpha_t}\vect{x}^{(i)}, \sigma_t^2 I) = \frac{\vect{x}_t-\sqrt{\alpha_t} \vect{x}^{(i)}}{\sigma_t}) \frac{\vect{x}_t-\sqrt{\alpha_t} \vect{x}^{(i)}}{\sigma_t} \\
        &= \sum_{\vect{x}^{(i)}\in D_y} \frac{p(\vect{x}^{(i)}| y)}{p(\vect{x}_t| y)}\frac{1}{(2\pi \sigma_t)^{\frac{n}{2}}} \exp{(-\frac{\|\vect{x}_t-\sqrt{\alpha_t}\vect{x}^{(i)}\|_2^2}{2 \sigma_t^2})} \frac{\vect{x}_t-\sqrt{\alpha_t} \vect{x}^{(i)}}{\sigma_t} \\
    \end{aligned}
\end{equation*}
To avoid numerical problem caused by $\frac{1}{(2\pi \sigma_t)^{\frac{n}{2}}}$ and intractable $\frac{p(\vect{x}^{(i)}| y)}{p(\vect{x}_t| y)}$, we re-organize this equation using softmax function:
\begin{equation*}
    \begin{aligned}
                \vect{\epsilon}_{\theta}(\vect{x}_t, t, y) 
                &= \sum_{\vect{x}^{(i)}\in D_y} \frac{\frac{p(\vect{x}^{(i)}| y)}{p(\vect{x}_t| y)}\frac{1}{(2\pi \sigma_t)^{\frac{n}{2}}} \exp{(-\frac{\|\vect{x}_t-\sqrt{\alpha_t}\vect{x}^{(i)}\|_2^2}{2 \sigma_t^2})}}{\sum_{j=1}^{|D_y|} p(\vect{x}_j|\vect{x}_t, y)} 
                \frac{\vect{x}_t-\sqrt{\alpha_t} \vect{x}^{(i)}}{\sigma_t} \\
                &= \sum_{\vect{x}^{(i)}\in D_y} \frac{\frac{p(\vect{x}^{(i)}| y)}{p(\vect{x}_t| y)}\frac{1}{(2\pi \sigma_t)^{\frac{n}{2}}} \exp{(-\frac{\|\vect{x}_t-\sqrt{\alpha_t}\vect{x}^{(i)}\|_2^2}{2 \sigma_t^2})}}{\sum_{j=1}^{|D_y|}\frac{p(\vect{x}_j| y)}{p(\vect{x}_t| y)}\frac{1}{(2\pi \sigma_t)^{\frac{n}{2}}} \exp{(-\frac{\|\vect{x}_t-\sqrt{\alpha_t}\vect{x}_j\|_2^2}{2 \sigma_t^2})}} \frac{\vect{x}_t-\sqrt{\alpha_t} \vect{x}^{(i)}}{\sigma_t} \\
                &=\sum_{\vect{x}^{(i)}\in D_y}\frac{1}{\sigma_t}(\vect{x}_t-\sqrt{\alpha_t}\vect{x}^{(i)})\frac{\exp(-\frac{1}{2\sigma_t^2}\|\vect{x}_t-\sqrt{\alpha_t}\vect{x}^{(i)}\|_2^2)}{\sum_{\vect{x}^{(j)} \in D_y} \exp(-\frac{1}{2\sigma_t^2}\|\vect{x}_t-\sqrt{\alpha_t}\vect{x}^{(j)}\|_2^2)}.
    \end{aligned}
\end{equation*}
This is the result of \cref{equation:optimal_diffusion}.
\end{proof}

\subsubsection{Optimal diffusion classifier: proof of \cref{theorem:3-3}}
\label{sec:optimal_diffusion_classifier}
\begin{proof}
Substitute \cref{equation:optimal_diffusion} into \cref{equation:diffusion_classifier}:
\begin{equation*}
\small
    \begin{aligned}
        &f_{\theta_D^*}(\vect{x})_y  \\
        =&-\mathbb{E}_{t,\vect{\epsilon}}[\|\vect{\epsilon}_{\theta}(\vect{x}_t, t, y)-\vect{\epsilon}\|_2^2] \\
=&-\mathbb{E}_{t, \vect{\epsilon}}[\|\sum_{\vect{x}^{(i)}\in D_y}[\frac{(\vect{x}_t-\sqrt{\alpha_t}\vect{x}^{(i)})}{\sigma_t}
\frac{\exp(-\frac{\|\vect{x}_t-\sqrt{\alpha_t}\vect{x}^{(i)}\|_2^2}{2\sigma_t^2})}
{\sum_{\vect{x}^{(j)} \in D_y} \exp(-\frac{\|\vect{x}_t-\sqrt{\alpha_t}\vect{x}^{(j)}\|_2^2}{2\sigma_t^2})}
]-\vect{\epsilon}\|_2^2] \\
=&-\mathbb{E}_{t, \vect{\epsilon}}[\|\sum_{\vect{x}^{(i)}\in D_y}[\frac{(\sqrt{\alpha_t}\vect{x}+\sigma_t\vect{\epsilon}-\sqrt{\alpha_t}\vect{x}^{(i)})}{\sigma_t}
\frac{\exp\left(-\frac{\|\sqrt{\alpha_t}\vect{x}+\sigma_t\vect{\epsilon}-\sqrt{\alpha_t}\vect{x}^{(i)}\|_2^2}{2\sigma_t^2}\right)}{
\sum_{\vect{x}^{(j)}\in D_y}
\exp\left(-\frac{\|\sqrt{\alpha_t}\vect{x}+\sigma_t\vect{\epsilon}-\sqrt{\alpha_t}\vect{x}^{(j)}\|_2^2}{2\sigma_t^2}\right)
}]
-\vect{\epsilon}\|_2^2]  \\
=&-\mathbb{E}_{t, \vect{\epsilon}}[\|\sum_{\vect{x}^{(i)}\in D_y}[\frac{1}{\sigma_t}(\sqrt{\alpha_t}\vect{x}-\sqrt{\alpha_t}\vect{x}^{(i)})s(\vect{x},\vect{x}^{(i)}, \vect{\epsilon}, t) + \vect{\epsilon} \sum_{\vect{x}^{(i)}\in D_y}s(\vect{x},\vect{x}^{(i)}, \vect{\epsilon}, t)]  \\
=&-\mathbb{E}_{t, \vect{\epsilon}}[\|\sum_{\vect{x}^{(i)}\in D_y}[\frac{1}{\sigma_t}(\sqrt{\alpha_t}\vect{x}-\sqrt{\alpha_t}\vect{x}^{(i)}) s(\vect{x},\vect{x}^{(i)}, \vect{\epsilon}, t)]\|_2^2]  \\
=&-\mathbb{E}_{\vect{\epsilon}, t}[\frac{\alpha_t}{\sigma_t^2}\|\sum_{\vect{x}^{(i)}\in D_y} s(\vect{x},\vect{x}^{(i)}, \vect{\epsilon}, t)(\vect{x}-\vect{x}^{(i)})\|_2^2].
    \end{aligned}
\end{equation*}
We get the result.
\end{proof}

%%%%%%%%%%%%%%%%%%%%%%%%%%%%%%%%%%%%%%%%%%%%%%%%%%%%%%%%%%%%%%%%%%%%%%%%%%%%%%%%%%%%%%%%%%%%%%%%%%%%%%%%%%%%%%%%%%%%%%%%%%%%%%%%%%%%%%%%%%%%%%%%%%%%%%%%%%%%

\subsection{Derivation of conditional elbos in \cref{equation:elbo_condition}}
We provide a derivation of conditional ELBO in the following, which is similar to the unconditional ELBO in \citet{ddpm}.

\begin{equation*}
    \begin{aligned}
        &\log p_{\theta}(\vect{x}_0|y)  \\
=&\log\int \frac{ p_{\theta}(\vect{x}_{0:T}|y) q(\vect{x}_{1:T}|\vect{x}_0,y)}{q(\vect{x}_{1:T}|\vect{x}_0,y)}d\vect{x}_{1:T} \\
=&\log \mathbb{E}_{q(\vect{x}_{1:T}|\vect{x}_0,y)}[\frac{ p_{\theta}(\vect{x}_{T}|y)p_{\theta}(\vect{x}_{0:{T-1}}|\vect{x}_{T},y)}{q(\vect{x}_{1:T}|\vect{x}_0,y)}] \\
\geq&  \mathbb{E}_{q(\vect{x}_{1:T}|\vect{x}_0,y)}[\log \frac{ p_{\theta}(\vect{x}_{T}|y)p_{\theta}(\vect{x}_{0:{T-1}}|\vect{x}_{T},y)}{q(\vect{x}_{1:T}|\vect{x}_0,y)}] \\
=&\mathbb{E}_{q(\vect{x}_{1:T}|\vect{x}_0,y)}[\log \frac{ p_{\theta}(\vect{x}_{T}|y)\prod_{i=0}^{T-1}p_{\theta}(\vect{x}_{i}|\vect{x}_{i+1},y)}{\prod_{i=0}^{T-1} q(\vect{x}_{i+1}|\vect{x}_i, \vect{x}_0,y)}] \\
=& \mathbb{E}_{q(\vect{x}_{1:T}|\vect{x}_0,y)}[\log \frac{ p_{\theta}(\vect{x}_{T}|y)\prod_{i=0}^{T-1}p_{\theta}(\vect{x}_{i}|\vect{x}_{i+1},y)}{\prod_{i=0}^{T-1} \frac{q(\vect{x}_{i+1}|\vect{x}_0,y)q(\vect{x}_{i}|\vect{x}_{i+1},\vect{x}_0,y)}{q(\vect{x}_i|\vect{x}_0,y)}}] \\
=&\mathbb{E}_{q(\vect{x}_{1:T}|\vect{x}_0,y)}[\log \frac{ p_{\theta}(\vect{x}_{T}|y)\prod_{i=0}^{T-1}p_{\theta}(\vect{x}_{i}|\vect{x}_{i+1},y)}{\prod_{i=0}^{T-1} q(\vect{x}_{i}|\vect{x}_{i+1},\vect{x}_0,y)}-\log q(\vect{x}_{T}|\vect{x}_0,y)] \\
=&\mathbb{E}_{q(\vect{x}_{1:T}|\vect{x}_0,y)}[\log \frac{\prod_{i=0}^{T-1}p_{\theta}(\vect{x}_{i}|\vect{x}_{i+1},y)}{\prod_{i=0}^{T-1} q(\vect{x}_{i}|\vect{x}_{i+1},\vect{x}_0,y)}-\log \frac{q(\vect{x}_{T}|\vect{x}_0,y)}{ p_{\theta}(\vect{x}_{T}|y)}] \\
=&\sum_{i=0}^{T-1}\mathbb{E}_{q(\vect{x}_i,\vect{x}_{i+1}|\vect{x}_0,y)}[\log \frac{p_{\theta}(\vect{x}_{i}|\vect{x}_{i+1},y)}{q(\vect{x}_{i}|\vect{x}_{i+1},\vect{x}_0,y)}]-D_{KL}(q(\vect{x}_{T}|\vect{x}_0,y)\| p_{\theta}(\vect{x}_{T}|y)) \\
=&\sum_{i=0}^{T-1}\mathbb{E}_{q(\vect{x}_{i+1}|\vect{x}_0,y)}\mathbb{E}_{q(\vect{x}_i|\vect{x}_{i+1},\vect{x}_0,y)}[\log \frac{p_{\theta}(\vect{x}_{i}|\vect{x}_{i+1},y)}{q(\vect{x}_{i}|\vect{x}_{i+1},\vect{x}_0,y)}]-D_{KL}(q(\vect{x}_{T}|\vect{x}_0,y)\| p_{\theta}(\vect{x}_{T}|y)) \\
=& C_4 -\sum_{i=1}^{T-1}\mathbb{E}_{q(\vect{x}_{i+1}|\vect{x}_0,y)}[D_{KL}(q(\vect{x}_i|\vect{x}_{i+1},\vect{x}_0,y)\|p_{\theta}(\vect{x}_{i}|\vect{x}_{i+1},y))] \\
=&  - \mathbb{\mathbb{E}}_{\vect{\epsilon}, t} \left[ w_t \|\boldsymbol{\mathbf{\epsilon}}_{\theta}(\vect{x}_t, t, y) - \boldsymbol{\mathbf{\epsilon}}\|_2^2 \right] + C.
    \end{aligned}
\end{equation*}
We get the result of \cref{equation:elbo_condition}.

%%%%%%%%%%%%%%%%%%%%%%%%%%%%%%%%%%%%%%%

% \subsection{Discussion}

\subsection{Connection between Energy-Based Models~(EBMs)}

The EBMs~\citep{lecun2006tutorial_ebm} directly use neural networks to learn $p_{\theta}(\vect{x})$ and $p_{\theta}(\vect{x}|y)$. 
\begin{equation*}
    p_{\theta}(\vect{x}|y) = \frac{\exp(-E_{\theta}(\vect{x})_y)}{Z(\theta, y)},
\end{equation*}
Where $E_{\theta}(\vect{x}) : R^D \to R^n$, and $Z(\theta, y) = \int \exp(-E_{\theta}(\vect{x})_y) d\vect{x}$ is the normalizing constant.

As described in \citet{grathwohl2019your_classifier_secret_ebm}, we can use EBMs to classify images by calculating the conditional probability:

\begin{equation}
\label{equation:ebm_classifier}
    p_{\theta}(y|\vect{x}) = \frac{\exp(-E_{\theta}(\vect{x})_y)}{\sum_{\hat{y}}\exp(-E_{\theta}(\vect{x})_{\hat{y}})}.
\end{equation}

Compare \cref{equation:ebm_classifier} and \cref{equation:diffusion_classifier}, we can also set the energy function as:

\begin{equation}
\label{equation:diffusion_classifier_is_ebm}
    E_{\theta}(\vect{x})_y \approx \mathbb{E}_{t, \vect{\epsilon}} \left[ w_t \|\vect{\epsilon}_{\theta}(\vect{x}_t, t, y) - \vect{\epsilon}\|_2^2 \right]. 
\end{equation}

Therefore, our diffusion classifier could be viewed as an EBM, and the energy function is the conditional diffusion loss.

\subsection{Computing gradient without computing UNet jacobi}

We propose another way to compute the gradient of \cref{equation:monte_carlo_t} without backpropagating through the UNet. Note that we do not use this method in any of the experiments. We only derive this method and conduct some theoretical analysis.
\begin{lemma}
    Assuming that $\vect{z} \sim \mathcal{N}(\vect{\mu}, \vect{\Sigma}),\; f:\mathbb{R}^{n_i} \to \mathbb{R}^{n_o} \in \mathcal{C}^1, \,\;p: \mathbb{R}^{n_i} \to \mathbb{R}  \in \mathcal{C}^1$. We can get
    \begin{equation}
        \nabla_{\vect{\mu}} \mathbb{E}_{\vect{z}}[f(\vect{z})] = \mathbb{E}_{\vect{z}}[\nabla_{\vect{\mu}}\log p(\vect{z})f(\vect{z})^T ].
    \end{equation}
    \label{theorem:high_dimension_NES}
    \vspace{-0.5cm}
\end{lemma}
\begin{proof}
    Inspired by \citet{wierstra2014natural}, we derive
    \begin{equation*}
        \begin{aligned}
            \nabla_{\vect{\mu}} E[f(\vect{z})] &=  \nabla_{\vect{\mu}}\int f(\vect{z})p(\vect{z}|\vect{\mu})d\vect{z} \\
&= \lim_{d \vect{\mu} \to \vect{0}} \frac{\int f(\vect{z})p(\vect{z}| \vect{\mu}+d \vect{\mu})d\vect{z} - \int f(\vect{z})p(\vect{z}|\vect{\mu})d\vect{z}}{d \vect{\mu}} \\
&=\int \nabla_{\vect{\mu}}p(\vect{z}|\vect{\mu})f(\vect{z})^Td\vect{z} \\
&=\int p(\vect{z}|\vect{\mu})\nabla_{\vect{\mu}}\log p(\vect{z}|\vect{\mu})f(\vect{z})^Td\vect{z} \\
&=\mathbb{E}_{z}[\nabla_{\vect{\mu}}\log p(\vect{z}|\vect{\mu})f(\vect{z})^T ].  \\
        \end{aligned}
    \end{equation*}
    \vspace{-0.5cm}
\end{proof}
According to Lemma \ref{theorem:high_dimension_NES}, we can derive the gradient of \cref{equation:monte_carlo_t} as
% \begin{equation}
%     \begin{aligned}
%         &\nabla_{\vect{x}} \mathbb{E}_{\vect{\epsilon}}[\|\vect{\epsilon}_{\theta}(\vect{x}_t,t)-\vect{\epsilon}\|_2^2]
%  \\
% = &2\mathbb{E}_{\vect{\epsilon}}[\Big(\nabla_{\vect{x}} \vect{\epsilon}_{\theta}(\vect{x}_t,t)\Big)  (\vect{\epsilon}_{\theta}(\vect{x}_t,t)-\vect{\epsilon})]  \\
% =&2\mathbb{E}_{\vect{\epsilon}}[ ( \nabla_{\vect{x}} \log p(\vect{\epsilon})\vect{\epsilon}_{\theta}(\vect{x}_t,t)^T (\vect{\epsilon}_{\theta}(\vect{x}_t,t)-\vect{\epsilon})]  \\
% =&2\mathbb{E}_{\vect{\epsilon}}[ ( \nabla_{\vect{x}} \log p(\vect{\epsilon}) \big( \vect{\epsilon}_{\theta}(\vect{x}_t,t)^T (\vect{\epsilon}_{\theta}(\vect{x}_t,t)-\vect{\epsilon}) \big)]. 
%     \end{aligned}
% \end{equation}

% \begin{equation}
%     \nabla_{\vect{x}} \mathbb{E}_{\vect{\epsilon}}[\|\vect{\epsilon}_{\theta}(\vect{x}_t,t)-\vect{\epsilon}\|_2^2]=2\mathbb{E}_{\vect{\epsilon}}[ \frac{\partial \log p(\vect{x}_t|\vect{x})}{\partial \vect{x}} \|\vect{\epsilon}_{\theta}(\vect{x}_t,t)-\vect{\epsilon}\|_2^2]. 
% \end{equation}

\begin{equation}
    \begin{aligned}
        &\frac{d}{d \vect{x}} \mathbb{E}_{\vect{\epsilon}}[\|\vect{\epsilon}_{\theta}(\vect{x}_t,t)-\vect{\epsilon}\|_2^2] \\
        =&\frac{d}{d \vect{x}} \mathbb{E}_{\vect{x}_t}[ \|\vect{\epsilon}_{\theta}(\vect{x}_t,t)-\frac{\vect{x}_t - \sqrt{\alpha_t}\vect{x}}{\sigma_t}\|_2^2] \\
        =&\frac{d}{d \vect{x}} \mathbb{E}_{\vect{x}_t}[g(\vect{x}_t, \vect{x}, t)] \\
        =&\frac{\partial }{\partial \vect{x}_t} \mathbb{E}_{\vect{x}_t}[g(\vect{x}_t, \vect{x}, t)] \frac{ \partial \vect{x}_t}{\partial \vect{x}} + \frac{\partial }{\partial \vect{x}} \mathbb{E}_{\vect{x}_t}[g(\vect{x}_t, \vect{x}, t)] \\
        =& \mathbb{E}_{\vect{x}_t}[\frac{\partial \log p(\vect{x}_t|\vect{x})}{\partial \vect{x}} g(\vect{x}_t, \vect{x}, t)] + \frac{\partial }{\partial \vect{x}} \mathbb{E}_{\vect{x}_t}[g(\vect{x}_t, \vect{x}, t)] \\
        =& \mathbb{E}_{\vect{x}_t}[\frac{\partial \log p(\vect{x}_t|\vect{x})}{\partial \vect{x}} \|\vect{\epsilon}_{\theta}(\vect{x}_t,t)-\frac{\vect{x}_t - \sqrt{\alpha_t}\vect{x}}{\sigma_t}\|_2^2] +  \mathbb{E}_{\vect{x}_t}[2(\vect{\epsilon}_{\theta}(\vect{x}_t,t)-\frac{\vect{x}_t - \sqrt{\alpha_t}\vect{x}}{\sigma_t}) \frac{\sqrt{\alpha_t}}{\sigma_t}] \\
        =&\mathbb{E}_{\vect{\epsilon}}[\frac{\partial \log p(\vect{x}_t|\vect{x})}{\partial \vect{x}} \|\vect{\epsilon}_{\theta}(\vect{x}_t,t)-\vect{\epsilon}\|_2^2] +  \mathbb{E}_{\vect{\epsilon}}[(\vect{\epsilon}_{\theta}(\vect{x}_t,t)-\vect{\epsilon})\frac{2\sqrt{\alpha_t}}{\sigma_t}].
    \end{aligned}
\end{equation}

Similarly, we can get the gradient of conditional diffusion loss

\begin{equation}
    \begin{aligned}
        &\frac{d}{d \vect{x}} \mathbb{E}_{\vect{\epsilon}}[\|\vect{\epsilon}_{\theta}(\vect{x}_t,t, y)-\vect{\epsilon}\|_2^2] \\
        =&\mathbb{E}_{\vect{x}_t}[\frac{\partial \log p(\vect{x}_t|\vect{x})}{\partial \vect{x}} \|\vect{\epsilon}_{\theta}(\vect{x}_t,t, y)-\frac{\vect{x}_t - \sqrt{\alpha_t}\vect{x}}{\sigma_t}\|_2^2] +  \mathbb{E}_{\vect{x}_t}[(\vect{\epsilon}_{\theta}(\vect{x}_t,t, y)-\frac{\vect{x}_t - \sqrt{\alpha_t}\vect{x}}{\sigma_t}) \frac{2\sqrt{\alpha_t}}{\sigma_t}] \\
        =&\mathbb{E}_{\vect{\epsilon}}[\frac{\partial \log p(\vect{x}_t|\vect{x})}{\partial \vect{x}} \|\vect{\epsilon}_{\theta}(\vect{x}_t,t, y)-\vect{\epsilon}\|_2^2] +  \mathbb{E}_{\vect{\epsilon}}[(\vect{\epsilon}_{\theta}(\vect{x}_t,t, y)-\vect{\epsilon})\frac{2\sqrt{\alpha_t}}{\sigma_t}].
    \end{aligned}
\end{equation}

As shown, the gradient of \cref{equation:monte_carlo_t} have two terms. The first term equals to the weighted sum of $\frac{\partial \log p(\vect{x}_t|\vect{x})}{\partial \vect{x}}$. In VE-SDE case, where $\vect{x}_t=\vect{x} + \sigma_t \vect{\epsilon}$, the negative gradient direction is aligned with $\vect{x}-\vect{x}_t$~(a vector starting from $\vect{x}_t$ and ending at $\vect{x}$). 
The second term is proportional to the gradient of Score Distillation Sampling~\citep{poole2022dreamfusion, wang2023prolificdreamer}, which also point toward real data.
Consequently, optimizing the diffusion loss will move $\vect{x}$ toward a region with higher log likelihood.

%%%%%%%%%%%%%%%%%%%%%%%%%%%%%%%%%%%%%%%%%%%%%%%%%%%%%%%%%%%%%%%%%%%%%%%%%%%%%%%%%%%%%%%%%%%%%%%%%%%%%%%%%%%%%%%%%%%%%%
\section{More experimental results}
\label{appendix:supplementary_exps_and_training}

\subsection{Training details}
\label{sec:training_details}

\textbf{Computational resources. }
We conduct Direct Attack on 1$\times$ A40 GPUs due to the large memory cost of computational graphs for second-order derivatives. We use 2$\times$ 3090 GPUs for other experiments. We also analyze the time complexity and test the real-time cost on a single 3090 GPU, as demonstrated in \cref{table:cifar_supplementary}. We are unable to assess the real-time cost of some methods due to difficulties in replicating them.

\textbf{Training details of multi-head diffusion. }
To reduce the time complexity of the diffusion classifier from $O(K\times T)$ to $O(T)$, we propose to slightly modify the architecture of the UNet, enabling it to predict for all classes at once. Since our changes are limited to the UNet architecture, all theorems and analyses remain applicable in this context.

However, this architecture only achieves 60\% accuracy on the CIFAR10 dataset, even with nearly the same number of parameters as the original UNet. We tried to solve this problem by using a larger CFG~(\textit{i.e.}, viewing extrapolated result $(1+\text{cfg})\cdot \vect{\epsilon}_{\theta}(\vect{x}_t,t, y)-\text{cfg} \cdot \vect{\epsilon}_{\theta}(\vect{x}_t,t)$ as the prediction of UNet), but it does not work.

\begin{algorithm}[t] %tb
\small
   \caption{Training of multi-head diffusion}
   \label{algorithm:multihead-diffusion}
\begin{algorithmic}[1]
   \REQUIRE
   A pre-trained diffusion model $\vect{\epsilon}_{\theta}$, dataset $\mathcal{D}$, a multi-head diffusion model $\vect{\epsilon}_{\phi}$
   \REPEAT
   \STATE $\vect{x}_0, y \sim D$;
   \STATE $t \sim $ Uniform($\{1,2,...,T\}$), $\vect{\epsilon} \sim \mathcal{N}(\vect{0}, \vect{I})$;
   \FOR{$y=0$ {\bfseries to} $K-1$}
   %\STATE \# first step
   \STATE Take gradient descent step on $\nabla_{\phi} \mathbb{E}_{\vect{\epsilon}, t}[w_t\|\vect{\epsilon}_\theta(\hat{\vect{x}}_t,t, y)-\vect{\epsilon}_\phi(\hat{\vect{x}}_t,t, y)\|_2^2]$;
   \ENDFOR
   \UNTIL converged;
% \STATE \textbf{Return:} $\tilde{y} = \argmax_y p_{\theta}(y|\vect{x})$.
\end{algorithmic}
\end{algorithm}

We hypothesize that with the traditional conditional architecture, the UNet focuses on extracting features relevant to specific class labels, leading to a more accurate prediction of the conditional score. In contrast, multi-head diffusion must extract features suitable for predicting all classes, as different heads use the same features for their predictions. To test this hypothesis, we measure the cosine similarity between features of a given $\vect{x}_t$ with different embeddings $y$. We find that for the traditional diffusion architecture, these features differ from each other. However, for multi-head diffusion, the cosine similarity of these predictions exceeds 0.98, indicating that the predictions are almost identical due to the identical feature.

It's worth noting that this does not mean traditional diffusion models are superior to multi-head diffusion. Both architectures have nearly the same number of parameters, as we only modify the last convolution layer. Additionally, the training loss curve and validation loss curve for both are almost identical, indicating they fit the training distribution and generalize to the data distribution similarly. The FID values of these two models are 3.14 and 3.13, very close to each other. The decreased performance of multi-head diffusion in the diffusion classifier is likely because it isn't clear on which feature to extract first. The training dynamic lets multi-head diffusion extracts features suitable for all classes, leading to similar predictions for each class, similar diffusion loss, and thus lower classification performance.

To prevent predictions for all classes from being too similar, we first considered training the multi-head diffusion with negative examples. Initially, we attempted to train the multi-head diffusion using the cross-entropy loss. While this achieved a training accuracy of 91.79\%, the test accuracy only reached 82.48\%. Moreover, as training continued, overfitting to the training set became more pronounced. \textbf{Notably, this model had 0\% robustness.} Fortunately, this experiment underscores the strength of our adaptive attacks in evaluating such randomized defenses, affirming that the robustness of the diffusion classifier is not merely due to its stochastic nature leading to an inadequate evaluation. A lingering concern is our lack of understanding as to why switching the training loss from diffusion loss to cross-entropy loss drastically diminishes the generalization ability and robustness.

Our hypothesis posits that, when trained with the diffusion loss, diffusion models are compelled to extract robust features because they are required to denoise the noisy images. However, when trained using the cross-entropy loss, there isn't a necessity to denoise the noisy images, so the models might not extract robust features. As a result, they may lose their image generation and denoising capabilities, as well as their generalization ability and robustness. We evaluated the diffusion loss of the diffusion models trained by cross-entropy loss and found that their diffusion losses hovered around 10. Furthermore, the images they generated resembled noise, meaning that they lose their generation ability.

To address this issue, we need to strike a balance between the diffusion loss, which ensures the robustness of the diffusion models, and the negative example loss (e.g., cross-entropy loss, CW loss, DLR loss) to prevent their predictions for various classes from becoming too similar. This balancing act turns the training of multi-head diffusion into a largely hyper-parameter tuning endeavor. To circumvent such a complex training process, we suggest distilling the multi-head diffusion from a pretrained traditional diffusion model. As illustrated in \cref{algorithm:multihead-diffusion}, the primary distinction between multi-head diffusion distillation and traditional diffusion model training is that the predictions for all classes provided by the multi-head diffusion model are simultaneously aligned with those of a pre-trained diffusion model.

Note that in \cref{algorithm:multihead-diffusion}, the predictions for different classes are computed in parallel. This approach sidesteps the need for tedious hyper-parameter tuning. Nevertheless, there's still potential for refinement. In this algorithm, the input pair $(\vect{x}_t, t, y)$ is not sampled based on its probability $p(\vect{x}_t, t, y) = \int p(\vect{x}|y) p(t) p(\vect{x}_t|\vect{x}) p(y) d\vect{x}$. This could be why the multi-head diffusion slightly underperforms compared to the traditional diffusion model. Addressing this issue might involve using importance sampling, a potential avenue for future research.

\begin{table}
\centering
\caption{Clean accuracy~(\%) and robust accuracy~(\%) of different methods against unseen threats.}
\setlength{\tabcolsep}{1.0pt}
\label{table:cifar_supplementary}
\begin{tabu}{l|c|cc|cccc} 
\toprule
\multirow{2}{*}{Method}                      & \multirow{2}{*}{Architecture} & \multirow{2}{*}{NFEs} & \multirow{2}{*}{Real Time (s)} & \multirow{2}{*}{Clean Acc} & \multicolumn{3}{c}{Robust Acc}                                                                   \\
                                             &                               &                       &                            &                            & $\ell_\infty$ norm      & $\ell_2$ norm           & Avg                                          \\ 
\hline
AT-DDPM-$\ell_\infty$                        & WRN70-16                      & $1$                   & 0.01                       & 88.87                      & 63.28                   & 64.65                   & 63.97                                        \\
AT-DDPM-$\ell_2$                             & WRN70-16                      & $1$                   & 0.01                       & 93.16                      & 49.41                   & 81.05                   & 65.23                                        \\
AT-EDM-$\ell_\infty$                         & WRN70-16                      & $1$                   & 0.01                       & 93.36                      & 70.90                   & 69.73                   & 70.32                                        \\
AT-EDM-$\ell_2$                              & WRN70-16                      & $1$                   & 0.01                       & 95.90                      & 53.32                   & 84.77                   & 69.05                                        \\
PAT-self                                     & AlexNet                       & $1$                   & 0.01                       & 75.59                      & 47.07                   & 64.06                   & 55.57                                        \\ 
\hline
DiffPure ($t^*=0.125$)                       & UNet                          & $126$                 & 0.72                       & 87.50                      & 40.62                   & 75.59                   & 58.11                                        \\
DiffPure ($t^*=0.1$)                         & UNet                          & $101$                 & 0.60                      & 90.97                      & 44.53                   & 72.65                   & 58.59                                        \\
SBGC                                         & UNet                          & $30TK$                &15.78                      & 95.04                      & 0.00                    & 0.00                    & 0.00                                         \\
HybViT                                       & ViT                           & $1$                   & 0.01                       & 95.90                      & 0.00                    & 0.00                    & 0.00                                         \\
JEM                                          & WRN28-10                      & $1$                   & 0.01                       & 92.90                      & 8.20                    & 26.37                   & 17.29                                        \\ 
\hline
\citet{perez2021enhancing}   & WRN70-16                      & $9$                   & n/a                        & 89.48                      & 72.66                   & 71.09                   & 71.87                                        \\
\citet{schwinn2022improving} & WRN70-16                      & $KN$                  & n/a                        & 90.77                      & 71.00                   & 72.87                   & 71.94                                        \\
\citet{blau2023classifier}   & WRN70-16                      & $KN$                  & n/a                        & 88.18                      & 72.02                   & 75.90                   & 73.96                                        \\ 
\hline
LM (ours)                                    & WRN70-16                      & $1+NT$                & 2.50                       & 95.04                      & 2.34                    & 12.5                    & 7.42                                         \\
LM (ours)                                    & WRN70-16                      & $1+N$                 & 0.10                       & 87.89                      & 71.68                   & 75.00                   & \textbf{73.34}                               \\
DC (ours)                                    & UNet                          & $TK$                  & 9.76                       & 93.55                      & 35.94                   & 76.95                   & \textbf{55.45}                               \\
RDC (ours)                                   & UNet                          & $NT+TK$               & 12.26                      & 93.16                      & 73.24                   & 80.27                   & \textbf{76.76}                               \\
RDC (ours)                                   & UNet                          & $N+TK$                & 9.86                       & 88.18                      & \textbf{80.07} & \textbf{84.76} & \textbf{82.42}  \\
RDC (ours)                                   & UNet                          & $N+T$                 & 1.43                       & 89.85                      & 75.67                   & 82.03                   & \textbf{78.85}                               \\
\bottomrule
\end{tabu}
\end{table}

\subsection{More Analysis and Discussion}
\label{appendix:more_exp}

\begin{wraptable}{r}{6cm}
\vspace{-1.9cm}
\centering
\caption{Gradient magnitudes.}
\label{table:gradient_magnitude}
\begin{tabu}{l|c}
\hline
Method & $\frac{1}{D}\|g\|_1$  \\
\hline
\citet{Engstrom2019Robustness}      & $7.7 \times 10^{-6}$                  \\
\citet{wong2020fast}     & $1.1 \times 10^{-5}$                  \\
\citet{salman2020adversarially}      & $6.6 \times 10^{-6}$                  \\
\citet{debenedetti2022light}      & $9.8 \times 10^{-6}$                  \\
Ours       & $8.2 \times 10^{-6}$ \\          
\hline
\end{tabu}
\vspace{-0.4cm}
\end{wraptable}

\textbf{Gradient magnitude. } When attacking the diffusion classifiers, we need to take the derivative of the diffusion loss. This process is similar to what is done when training diffusion models, so gradient vanishing is unlikely to occur. We also measure the average absolute value of the gradient (\textit{i.e.}, \( \frac{1}{D}\|g\|_1 \)). As shown in \cref{table:gradient_magnitude}, the magnitude of the gradient in our method is on the same scale as that of other adversarial training models, validating that our method does not suffer from gradient vanishing.

\textbf{Substituting likelihood maximization with DiffPure.}
We further study the performance by substituting likelihood maximization with DiffPure. We use the same hyperparameters as in \citet{nie2022diffpure} and follow the identical evaluation setup as described in \cref{sec:4-1}. The robustness of each method under the $\ell_\infty$-norm threat model with $\epsilon_\infty = 8/255$ on the CIFAR-10 dataset is shown in \cref{tab:diffpure+dc}. As shown, DC+DiffPure outperforms DiffPure significantly, highlighting the effectiveness of our diffusion classifier. Furthermore, RDC surpasses DC+DiffPure, indicating that likelihood maximization is more compatible with the diffusion classifier. Besides, \citet{xiao2022densepure} provide an interesting explanation of DiffPure. It has been demonstrated that DiffPure increases the likelihood of inputs with high probability, resulting in better robustness. By directly maximizing the likelihood of inputs, our likelihood maximization further enhances the potential for improved robustness.

\begin{wraptable}{r}{5cm}
\centering
\vspace{-0.6cm}
\caption{The robustness of DiffPure, DiffPure+DC and RDC.}
    \begin{tabular}{l|c}
    \toprule
      Method   & Robustness(\%) \\
    \midrule
       DiffPure  & 53.52 \\
       DiffPure+DC & 69.92 \\
       RDC & 75.67 \\
    \bottomrule
    \end{tabular}
    \label{tab:diffpure+dc}
\vspace{-0.3cm}
\end{wraptable}

\textbf{Attacking using the adaptive attack in \citet{sabour2015adversarial}. }
\citet{tramer2020adaptive} propose to add an additional feature loss~\citep{sabour2015adversarial} that minimizes the class score between the current image and a target image in another class. This create adversarial examples whose class scores match those of clean examples but belong to a different class, thereby generating in-manifold adversarial examples, avoiding to be detected by likelihood-based adversarial example detectors. 
To evaluate the robustness of our method against these adaptive attacks, we integrate them with AutoAttack and test the robust accuracy under $\ell_\infty$ threat model with $\epsilon_\infty=8/255$. Surprisingly, our method achieves 90.04\% robustness against attack using feature loss, and 86.72\% robustness against attack using feature loss combined with the cross entropy loss or DLR loss in AutoAttack. 
On one hand, our Lagrange attack in \cref{sec:adaptive_attack} directly maximizes the lower bound of likelihood, making it more effective than feature loss. On the other hand, our method does not incorporate adversarial example detectors, making it unnecessary to strictly align the logits of adversarial examples with those of clean images.

\textbf{Comparison with other dynamic defenses. }
We also compare our methods with state-of-the-art dynamic defenses. As some of these methods have not yet been open-sourced, we reference the best results reported in their respective papers. We use $N$ to denote the optimization steps in their methods (\textit{e.g.}, qualification steps in \citet{schwinn2022improving}, PGD steps in \citet{blau2023classifier}). As shown in \cref{table:cifar_supplementary}, our methods are not only more efficient but also effective than these dynamic defenses. Specifically, the time complexities of these dynamic defenses are related to the number of classes $K$, which limits their applicability in large datasets. On the contrary, the time complexity of our RDC does not depend on $K$. Moreover, our RDC outperforms previous methods by +3.01\% on $\ell_\infty$ robustness and +6.33\% on $\ell_2$ robustness, demonstrating the strong efficacy and efficiency of our RDC.

\begin{wraptable}{r}{8cm}
\vspace{-0.5cm}
    \centering
    \caption{Comparison with other randomized defenses.}
    \begin{tabular}{l|cc}
    \toprule
      Method   & Attacker & Robustness(\%) \\
    \midrule
      \citet{fu2021double}   &  PGD-100 & 66.28 \\
      \citet{dong2022random} & PGD-20 & 60.69 \\
      \citet{hao2022gsmooth} & n/a & 0 \\
      RDC (Ours) & AutoAttack & 75.67 \\
      \bottomrule
    \end{tabular}
    \label{tab:compare_with_randomization_defense}
    \vspace{-0.3cm}
\end{wraptable}

\textbf{Comparison with other randomized defenses. }
As shown in \cref{tab:compare_with_randomization_defense}, our method outperforms previous state-of-the-art randomized defenses. This is because diffusion models are naturally robust to such Gaussian corruptions, and such high variance Gaussian corruptions are much more effective than \citet{fu2021double, dong2022random} to smooth the local extrema in loss landscape, preventing the existence of adversarial examples. Our method can also be integrated with randomized smoothing to get certified robustness. For more detail, refer to \citet{chen2024your}.

\textbf{Comparison between different likelihood maximizations. }
We compare the LM~($1+NT$) with the improved version LM~($1+N$). Surprisingly, under the BPDA attack, LM~($1+NT$) achieves only 2.34\% robustness. 
On the one hand, the likelihood maximization moves the inputs towards high log-likelihood region estimated by diffusion models, instead of traditional classifiers, thus it is more effective when combined with diffusion classifiers. On the other hand, although the diffusion losses of LM~($1+NT$) and LM~($1+N$) are same in expectation, the former induces less randomness, thus it is less effective to smooth the local extrema. LLet's delve into a special case with $N=1$. In this case, the expectation of LM~($1+NT$) is $\mathbb{E}_{\vect{\epsilon}}[f(\vect{x}+\nabla_{\vect{x}} \mathbb{E}_{t}[w_t\|\vect{\epsilon}_\theta(\vect{x}_t,t,y)-\vect{\epsilon}\|_2^2])]$, while the expectation of LM~($1+N$) is $\mathbb{E}_{\vect{\epsilon}, t}[f(\vect{x}+\nabla_{\vect{x}} [w_t\|\vect{\epsilon}_\theta(\vect{x}_t,t,y)-\vect{\epsilon}\|_2^2])]$. The primary difference between these two is the placement of the expectation over $T$ for LM~($1+N$), which is outside the function $f$. This arrangement implies that the randomness associated with $t$ also aids in smoothing out local extrema, leading to better smoothed landscape and higher robustness. It is essential to clarify that this is not a result of the stochasticity hindering the evaluation of their robustness. We have already accounted for their stochasticity by applying EOT 100 times, as illustrated in \cref{fig:randomness}. Note that the likelihood maximization acts like a pre-processing module, and it can be used to defend against adversarial attacks to any models in a plug-and-play manner, including current threat models toward large vision-language models~\citep{wei2023jailbreak,dong2023robust}.

% This result suggests that the BPDA attack is effective against optimization-based defenses, and the robustness of RDC is not attributed to the indifferentiability of likelihood maximization.

\textbf{Comparison between different RDCs. } 
As shown in \cref{table:cifar_supplementary}, our vanilla RDC attains 73.24\% $\ell_\infty$ robustness and 80.27\% $\ell_2$ robustness, surpassing prior adversarial training and diffusion-based purification techniques. By substituting the LM with the enhanced likelihood maximization, we manage to further boost the robustness by 6.83\% and 4.49\% against the $\ell_\infty$ and $\ell_2$ threat models, respectively. When employing multi-head diffusion, the RDC's time complexity significantly diminishes, yet its robustness and accuracy remain intact. This underscores the remarkable efficacy and efficiency of our proposed RDC.

\begin{table}[h]
    \centering
    \caption{Performance of DiffPure when using different architectures and checkpoints.}
    \begin{tabular}{c|c|c|c|c|c}
        \hline
        Method & Architecture & Diffusion model & Clean Acc & \multicolumn{2}{c}{Robust Acc} \\
        \cline{5-6}
         &  &  &  & $\ell_\infty$ norm & $\ell_2$ norm \\
        \hline
        DiffPure & UNet+WRN-70-16 & Score-SDE & 90.97\% & 43.75\% & 55.47\% \\
        DiffPure & UNet+WRN-70-16 & EDM & 92.58\% & 42.27\% & 60.94\% \\
        RDC & UNet & EDM & 89.85\% & 75.67\% & 82.03\% \\
        \hline
    \end{tabular}
    \label{tab:different_diffusion}
\end{table}

\textbf{Ablation studies on diffusion checkpoints.} We also implemented DiffPure using EDM checkpoints, decoupling the selection of checkpoints and samplers. One can use EDM checkpoints with EDM samplers or previous DDIM/DDPM samplers. The code can be found in our repository mentioned in the abstract. As shown in \cref{tab:different_diffusion}, when using EDM checkpoints, there is a slight improvement in clean accuracy and \(\ell_2\) robustness. However, DiffPure still lags significantly behind our method, as diffusion models still cannot completely purify the adversarial perturbations for the subsequent discriminative classifiers.

\textbf{Attacks using multiple EOT steps. } Since our methods only induce a small randomness on the gradient (see \cref{fig:randomness}), the Expectation Over Transformations (EOT) does not help when attacking our defense. We evaluate the robustness on the first 64 samples of the CIFAR-10 test set against the \(\ell_\infty\) threat model with \(\epsilon_\infty = 8/255\), using EOT numbers of 1, 5, and 10, respectively. Under all evaluations, RDC achieve 68.75\% robustness.

% \textbf{Momentum optimizer.} \huanran{Now we use adam. If we have time, we can redo this. }We then study the optimization algorithm for likelihood maximization in \cref{equation:optimization_real}. Optimizing without momentum results in 69\% robustness, which is approximately 7\% lower than that with momentum. This suggests that a better optimization algorithm could decrease the unconditional diffusion loss more effectively, thus improving the robustness. Using more advanced optimization algorithms for likelihood maximization may lead to even greater robustness, and we leave this to future works.

% ------------------------------------------------------------------------------------------------------------
\subsection{Experiment on Restricted ImageNet}

\textbf{Datasets and training details.} We conduct additional experiments on Restricted ImageNet~\citep{tsipras2019robustness}, since \citet{karras2022elucidating} provides off-the-shelf conditional diffusion model for imagenet dataset. Restricted ImageNet is a subset of ImageNet with 9 super-classes. 
For robustness evaluation, we randomly select 256 images from Restricted ImageNet test set due to the high computational cost of the attack algorithms, following \citet{nie2022diffpure}.

\textbf{Hyperparameters and robustness evaluation.} We use the same hyper-parameters and robustness evaluation as in \cref{sec:4-1}. Following common settings in adversarial attacks \citep{wong2020fast,nie2022diffpure,zhang2024duality}, we only evaluate $\ell_\infty$ robustness with $\epsilon_\infty=4/255$ in this subsection.

\textbf{Compared methods.} We compared our method with four state-of-the-art adversarial training models~\citep{Engstrom2019Robustness, wong2020fast, salman2020adversarially, debenedetti2022light,wei2023cfa} and DiffPure~\citep{nie2022diffpure}. For discriminative classifiers such as adversarially trained models, DiffPure, and LM, we compute the logit for each super-class by averaging the logits of its associated classes. 
For our RDC, we select the logit of the first class within the super-class to stand for the whole super-class.

\textbf{Results. } As shown in \cref{tab:restricted_imagenet}, our RDC outperforms previous methods by +1.75\%, even though RDC only uses the logit of the first class of each super class for classification. This demonstrates that our method is effective on other datasets as well.

\begin{table}[t]
    \centering
    \begin{tabular}{cc}
        \begin{minipage}[t]{0.45\textwidth}
            \centering
            \caption{Clean Accuracy (\%) and robust accuracy (\%) on CIFAR-100.}
            \label{table:cifar100}
            \begin{tabu}{l|cc} 
            \toprule
             Method      &  Clean Acc  &  Robust Acc  \\ 
            \midrule
            WRN40-2              & 78.13               & 0.00                  \\
            \citet{rebuffi2021fixing_data_aug_improve_at} & 63.56                & 34.64                 \\
            \citet{wang2023better_diffusion_improve_AT}   & 75.22                & 42.67                 \\
            DiffPure              & 39.06                & 7.81                   \\
            DC                    & 79.69                & 39.06                 \\
            RDC                   & \textbf{80.47}                & \textbf{53.12}                 \\
            \bottomrule
            \end{tabu}
        \end{minipage}
        &
        \begin{minipage}[t]{0.45\textwidth}
            \centering
            \caption{Clean accuracy~(\%) and robust accuracy~(\%) of different methods in Restricted ImageNet.}
            \label{tab:restricted_imagenet}
            \begin{tabular}{l|cc}
            \toprule
               Method  & Clean Acc  & Robust Acc \\
               \midrule
               \citet{Engstrom2019Robustness}  & 87.11 & 53.12 \\
               \citet{wong2020fast} & 83.98 & 46.88 \\
               \citet{salman2020adversarially} & 86.72 & 56.64 \\
               \citet{debenedetti2022light} & 80.08 & 38.67 \\
               DiffPure \citep{nie2022diffpure} & 81.25 & 29.30 \\
               % LM~(ours) & 85.16 & 69.53 \\
               RDC~(ours) & \bf87.50 &  \bf58.40    \\
               \bottomrule
            \end{tabular}
        \end{minipage}
    \end{tabular}
\end{table}

\subsection{Experiment on CIFAR-100}

We also test the robustness of different method against $\ell_\infty$ threat model with $\epsilon_\infty=8/255$, following the same experimental settings as CIFAR-10. Due to the time limit, we only random sample 128 images. The results are shown in \cref{table:cifar100}.

% \begin{table}
% \centering
% \caption{Clean Accuracy (\%) and robust accuracy (\%) on CIFAR-100.}
% \label{table:cifar100}
% \begin{tabu}{l|cc} 
% \toprule
%  Method      &  Clean Acc  &  Robust Acc  \\ 
% \midrule
% WRN40-2              & 78.13               & 0.00                  \\
% \citet{rebuffi2021fixing_data_aug_improve_at} & 63.56                & 34.64                 \\
% \citet{wang2023better_diffusion_improve_AT}   & 75.22                & 42.67                 \\
% DiffPure              & 39.06                & 7.81                   \\
% DC                    & 79.69                & 39.06                 \\
% RDC                   & \textbf{80.47}                & \textbf{53.12}                 \\
% \bottomrule
% \end{tabu}
% \end{table}

We find that RDC still achieves superior result compared with the state-of-the-art adversarially trained models and DiffPure.
More surprisingly, we discover that DiffPure does not work well on CIFAR-100. We guess this is because CIFAR-100 has more fine-grained classes, and thus a small amount of noise will make the image lose its semantic information of a specific class. Hence, DiffPure is not suitable for datasets with more fine-grained classes but small resolution. This experiment indicate that our methods could be easily scaled to fine-grained datasets.

\subsection{Discussions.}

\begin{figure}[t]
\centering
\subfigure[ELBO on CIFAR10-C]{
\includegraphics[width=8cm]{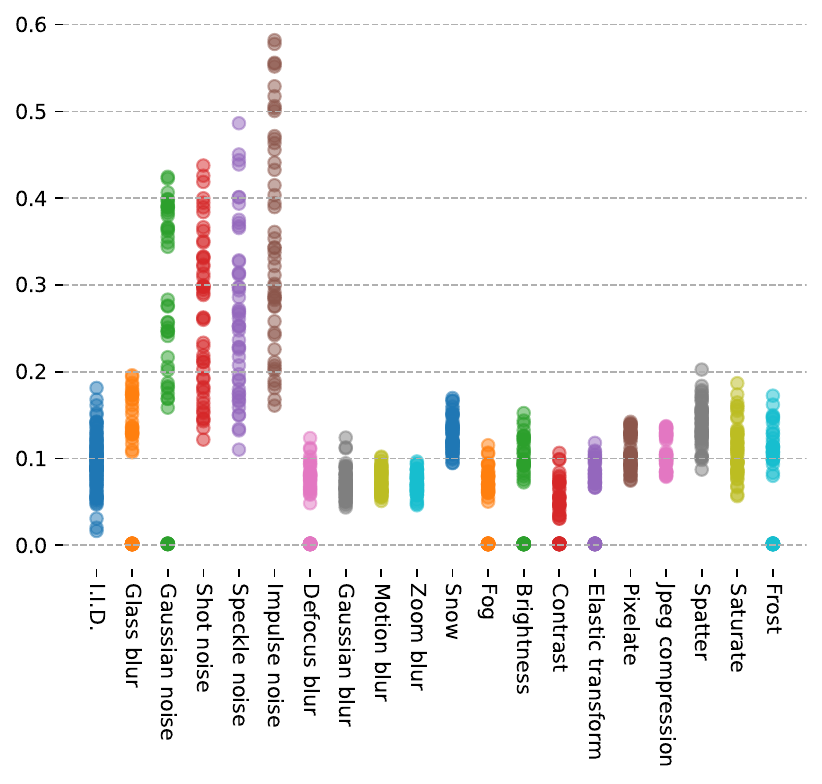}
\label{fig:ood_elbo}
}
\subfigure[BPD on CIFAR10-C]{
\includegraphics[width=8cm]{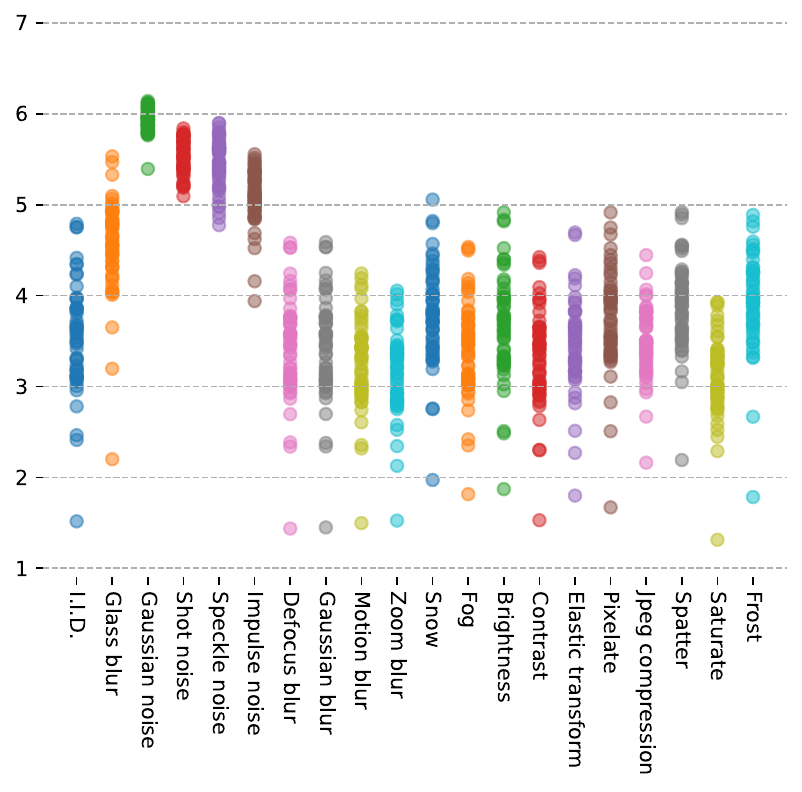}
\label{fig:ood_likelihood}
}
% \vspace{-2ex}
\caption{The prediction of ELBO and BPD on CIFAR-10 test set and CIFAR-10-C.}   
\label{fig:ood_detection}
\end{figure}

\textbf{O.O.D. Detection.} We test both the unconditional ELBO and the likelihood (expressed in Bits Per Dim (BPD) as mentioned in \citet{papamakarios2017masked}). We evaluate these metrics on the CIFAR-10 test set and CIFAR10-C. As demonstrated in \cref{fig:ood_detection}, while both methods can distinguish in-distribution data from certain types of corruptions, such as Gaussian blur and Gaussian noise, they struggle to differentiate in-distribution data from corruptions like fog and frost.

\begin{figure}[t]
    \centering
    \scalebox{0.57}{
    \includegraphics{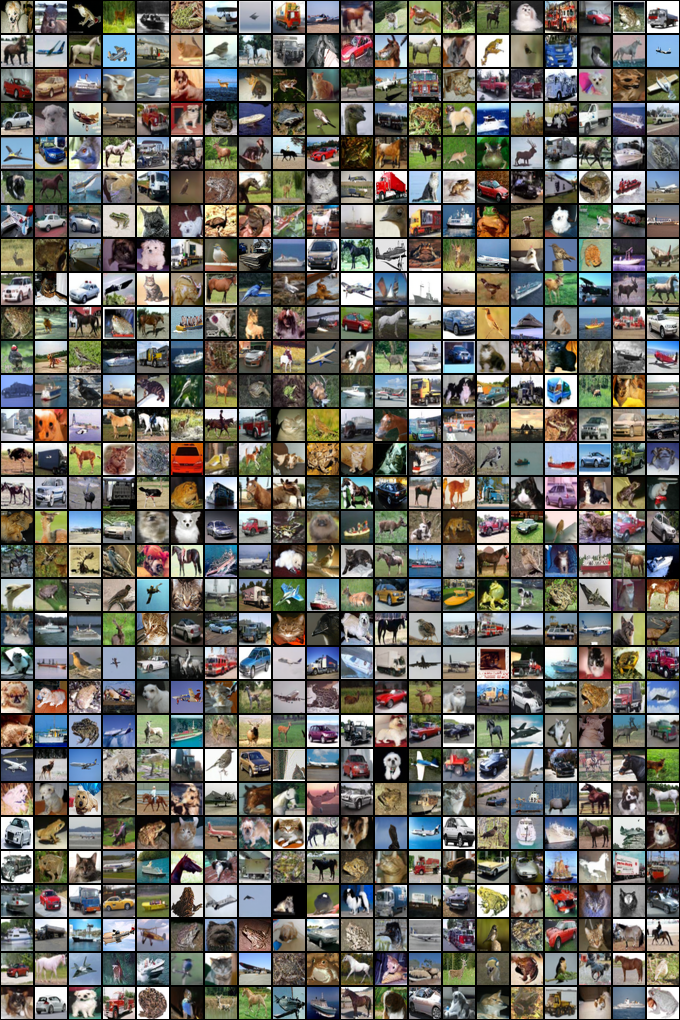}
    }
    \caption{The images generated by multi-head diffusion.}
    \label{fig:multi-head}
\end{figure}

\textbf{Generation of multi-head diffusion.} Since our multi-head diffusion is initialized from an unconditional EDM and distilled by a conditional EDM, it achieves a generative ability comparable to EDM. The images generated by our multi-head diffusion are shown in \cref{fig:multi-head}.

% -----------------------------------------------------------------------------------------------------
\section{Limitations}

Despite the great improvement, our methods could still be further improved. Currently, our methods requires $N+T$ NFEs for a single images, and applying more efficient diffusion generative models~\citep{song2023consistency_model, shao2023catch, liu2023instaflow} may further reduce $T$. Additionally, while we directly adopt off-the-shelf diffusion models from \citet{karras2022elucidating}, designing diffusion models specifically for classification may further improve performance. We hope our work serves as an encouraging step toward designing robust classifiers using generative models.

\end{document}